\documentclass[10pt]{article}

\usepackage[utf8]{inputenc}
\usepackage[T1]{fontenc}
\usepackage{amsmath,amssymb,amsthm}
\usepackage{booktabs}
\usepackage{multirow}
\usepackage[bookmarks=false]{hyperref}
\usepackage{cleveref}
\usepackage{algorithm}
\usepackage{algorithmic}
\usepackage{graphicx}
\usepackage{microtype}
\usepackage{xfrac}
\usepackage[margin=1in]{geometry}
\usepackage{natbib}
\usepackage{enumitem}
\usepackage{tikz}
\usetikzlibrary{arrows.meta, positioning, shapes.geometric, calc, decorations.pathreplacing, fit}
\usepackage{pgfplots}
\pgfplotsset{compat=1.18}

\newtheorem{definition}{Definition}

\newcommand{\clamp}{\operatorname{clamp}}
\newcommand{\softmax}{\operatorname{softmax}}
\DeclareMathOperator*{\argtopK}{arg\,top\text{-}K}

\title{Ailed: A Psyche-Driven Chess Engine\\with Dynamic Emotional Modulation}

\author{
  Diego Armando Resendez Prado\\
  Independent Researcher\\
  \texttt{diego.resendez@zero-oneit.com}
}

\date{}

\begin{document}

\maketitle

% ============================================================================
% ABSTRACT
% ============================================================================
\begin{abstract}
Chess engines passed human strength years ago, but they still don't play like humans. A grandmaster under clock pressure blunders in ways a club player on a hot streak never would. Conventional engines capture none of this.

This paper proposes a \textbf{personality $\times$ psyche decomposition} for producing behavioral variability in chess play, taking cues from patterns seen in human games. \textbf{Personality} is static---a preset that pins down the engine's character. \textbf{Psyche} is dynamic---a bounded scalar $\psi_t \in [-100, +100]$, recomputed from five positional factors after every move. These two components feed into an \textbf{audio-inspired signal chain} (noise gate, compressor/expander, five-band equalizer, saturation limiter) that reshapes move probability distributions on the fly. The chain doesn't care what engine sits behind it: any system that outputs move probabilities will do. It needs no search and carries no state beyond~$\psi_t$.

I test the framework across 12{,}414~games against Maia2-1100, feeding it two probability sources that differ by ${\sim}$2{,}800$\times$ in training data. Both show the same monotonic gradient in top-move agreement (${\sim}$20--25~pp spread from stress to overconfidence), which tells us the behavioral variation comes from the signal chain, not from the model underneath. When the psyche runs overconfident, the chain mostly gets out of the way (66\% agreement with vanilla Maia2). Under stress, competitive score falls from 50.8\% to 30.1\%. The patterns are reminiscent of tilt and overconfidence as described in human play, but I should be upfront: this study includes no human-subject validation.
\end{abstract}

% ============================================================================
% ============================================================================
% 1. INTRODUCTION
% ============================================================================
\section{Introduction}

Between alpha-beta search \citep{campbell2002deep} and reinforcement learning \citep{silver2018general, schrittwieser2020mastering}, chess engines have long since left human players behind. But as opponents, they share a basic limitation: they don't play like humans do. A grandmaster under clock pressure blunders. A club player on a roll takes risks that make no positional sense. A serious player goes home, reviews the loss, and comes back sharper. Stress degrading performance, confidence breeding recklessness, deliberate study paying off later---none of this shows up in conventional engines.

My starting point is simple: if we model dynamics \emph{drawn from} emotional patterns in decision-making---rather than treating all variability as noise---we can build chess opponents that are more engaging and far less predictable. Prior work on human-like play has mostly gone after matching move distributions at fixed skill levels \citep{mcilroy2020aligning}, or bolting fixed personality biases onto evaluation functions \citep{hudlicka2009affective}. Both approaches are static. They tell you what a human at a given level \emph{would} play, but not \emph{why} that same player deviates from their own baseline as the game goes on.

This paper introduces the \textbf{psyche model} paired with an \textbf{audio-inspired signal chain}---a dynamic framework, informed by emotional dynamics, that rests on three formal components:

\begin{enumerate}
    \item A \textbf{psyche state} $\psi_t \in [-100, +100]$ computed from five weighted positional factors after each move~$t$, representing the engine's computed reaction to board changes.
    \item A \textbf{signal chain} that reshapes the move probability distribution through seven stages---noise gate, dynamics (compressor/expander), five-band equalizer, and saturation---all parameterized by $\psi_t$ and a static personality preset. The chain is a pure probability transform: no search, no game history.
    \item A \textbf{temporal dynamics model} with within-game persistence and overnight exponential decay $\psi_{t+1}^{\text{day}} = (1 - \lambda)\psi_t^{\text{day}}$, giving rise to patterns such as tilt-like cascades and gradual recovery.
\end{enumerate}

I build the psyche model into \textbf{Ailed}, a chess engine powered by a 23.7M-parameter decoder-only transformer. Beyond the signal chain (which is the primary contribution and the focus of this paper's experiments), Ailed includes two additional psyche-aware modules: a lookahead planner whose plans can shatter under psyche pressure (\Cref{sec:thinking}), and an offline study system whose quality degrades when the psyche is at extremes (\Cref{sec:study}).

\paragraph{Scope.} This is not a chess engine in the usual sense, and shouldn't be evaluated as one. Stockfish, Leela Chess Zero, and AlphaZero all optimize for move quality---minimizing centipawn loss---which is a fundamentally different goal from behavioral realism. Where those engines are designed to eliminate cognitive limitations, this system deliberately puts them in. No deep search, no opening books, no endgame tablebases, no tactical solvers. That isn't a gap in the implementation; it's the point. The system has three purposes: generating play that shows imprecision and pressure sensitivity echoing patterns seen in human chess; giving users a training partner whose mistakes adapt to its psyche state, and providing middleware that can sit on top of any engine that outputs move probabilities.

The experiments (\Cref{sec:experiments}) pit the signal chain against Maia2-1100 across 12{,}414~games, using two different probability sources: Ailed's own model and Maia2's probabilities fed in from outside. In short: (i)~the chain creates a clean monotonic gradient in top-move agreement (${\sim}$20--25~pp spread), (ii)~the gradient holds up regardless of whether the probability source is weak or strong, (iii)~overconfident play degenerates into a near-transparent pass-through of the underlying model, and (iv)~stress drags down competitive score in a measurable way.

\paragraph{Contributions.} This paper contributes:
\begin{enumerate}[leftmargin=2em]
  \item A \textbf{formal personality $\times$ psyche decomposition}: personality as a static preset, psyche as a dynamic bounded scalar, behavior as their product.
  \item An \textbf{audio-inspired signal chain} (gate, dynamics, EQ, saturation) for psyche-modulated move selection that operates as architecture-agnostic middleware on any probability distribution.
  \item \textbf{Six personality presets} inspired by music genres, illustrating that static character and dynamic mood are orthogonal design dimensions.
  \item \textbf{Two cognitive extensions} (thinking mode and study mode) that share the psyche state, demonstrating its generality as a modulation mechanism.
  \item \textbf{Empirical validation} through 12{,}414~games across two engines and four experiments, including stage-wise ablation identifying the dynamics stage as the dominant behavioral mechanism.
  \item An \textbf{open-source implementation} including the signal chain, match runner, psyche replay, and analysis tools.
\end{enumerate}

% ============================================================================
% 2. RELATED WORK
% ============================================================================
\section{Related Work}

\paragraph{Neural chess engines.}
AlphaZero \citep{silver2018general} proved that a self-play-trained neural network could leapfrog decades of handcrafted engine work, pairing deep convolutional networks with Monte Carlo Tree Search. Leela Chess Zero later brought a similar approach to the open-source community. Both aim squarely at optimal play. What I do here is different: train on actual human games (1025--1175 ELO, from Lichess), learn what those players tend to play, and then push and pull those distributions through a dynamic psyche variable.

\paragraph{Human-like game AI.}
Maia Chess \citep{mcilroy2020aligning} took a different tack: train ELO-specific models to predict what a human would actually play, and it got quite good at it. Maia2 \citep{tang2024maia2} goes further with ELO-aware attention, so a single model handles any skill level. McIlroy-Young et al.~\citep{mcilroyyoung2022individual} then added per-player fine-tuning to capture individual style. Their emphasis is personalization across players; mine is nonstationarity \emph{within} a single game. Shoresh and Loewenstein~\citep{shoresh2024centaur} propose a Centaur that blends Maia (the human clone) with Leela (the strong engine) through a mixture-of-experts. Like Ailed, it treats a human-like model as middleware---but the blend is fixed. The twist in this work is making that layer \emph{dynamic}: the same engine plays differently depending on what just happened on the board. I compare directly against Maia2 in \Cref{sec:experiments}.

\paragraph{Affective computing in games.}
Most emotion modeling in games has gone in one of two directions: detecting what the player is feeling \citep{yannakakis2011experience}, or making NPCs that respond to it \citep{hudlicka2009affective}. The OCC model \citep{ortony1988occ} generates emotions from event appraisal; EMA \citep{marsella2009ema} turns that into a dynamic process for believable agents. But the common thread is that these systems model the \emph{human's} state and react to it. I flip this around. I give the \emph{AI itself} a dynamic internal state, informed by affect models, and let that state degrade its own performance. The closest precedent I'm aware of is work on mood in RL agents \citep{moerland2018emotion}, where mood modulates reward sensitivity. Rutledge et al.~\citep{rutledge2014wellbeing} gave me an empirical anchor for scalar affect---their reward-prediction-error model of momentary well-being informed how I designed $\psi$ as an accumulated position-evaluation signal. Where the psyche model parts ways with these is in \emph{what} it modulates: not reward, but the output distribution itself, through a multi-stage signal chain that leaves the learned policy intact while changing how it executes.

\paragraph{Probability distribution shaping.}
Temperature scaling is the standard knob for controlling output diversity in language models \citep{holtzman2020curious}. More recent work has pushed beyond that single parameter: Mirostat \citep{basu2020mirostat} runs a perplexity feedback loop, min-p \citep{nguyen2025minp} truncates relative to the top token, conformal nucleus sampling \citep{ravfogel2023conformal} ties nucleus size to entropy, and locally typical sampling \citep{meister2023locally} replaces probability rank with typicality. All of these optimize for text quality or diversity. What I'm after is different: the Ailed signal chain is parameterized by an affective state tied to chess features, and it targets behavioral realism---getting patterns that look like tilt and overconfidence---not output quality per se. Rather than one temperature parameter, I use a multi-stage transform. Each stage tackles a different facet of bounded decision-making: how wide the attention is (gate), how concentrated the choice becomes (dynamics), how different quality tiers get weighted (EQ), and whether any single move can dominate (saturation).

On the theoretical side, information-theoretic bounded rationality \citep{ortega2015bounded} gives a principled basis for softmax-like policies under resource constraints, quantal response equilibria \citep{mckelveypalf1995qre} frame temperature-like choice as rationality noise, and prospect theory \citep{kahneman1979prospect} supports reading stress and overconfidence as shifts in risk preference. I don't derive Ailed from free-energy optimality. It's a deliberately interpretable, domain-grounded transform stack---built to sit as middleware on top of whatever probability source you have.

% 3. PRELIMINARIES AND NOTATION
% ============================================================================
\section{Preliminaries and Notation}\label{sec:notation}

I define the formal objects used throughout the paper.

\paragraph{Board state.} Let $\mathcal{B}$ denote the set of legal chess board positions. A game is a sequence of positions $b_0, b_1, \ldots, b_N$ where $b_0$ is the standard starting position and each $b_{t+1}$ results from applying a legal move to $b_t$.

\paragraph{Move space.} Let $\mathcal{M}$ denote the set of all syntactically valid UCI moves. Moves are encoded using a fixed vocabulary $\mathcal{V}$ of $|\mathcal{V}| = 4{,}547$ tokens: 3~special tokens (\texttt{<pad>}, \texttt{<start>}, \texttt{<end>}), 4,032 standard moves (all source-destination pairs), and 512 promotion moves (with piece suffix $\in \{q, r, b, n\}$). At position $b_t$, the set of legal moves $\mathcal{L}(b_t) \subset \mathcal{M}$ is determined by chess rules.

\paragraph{Move history.} A move history up to time $t$ is a token sequence $\mathbf{x}_{1:t} = (x_1, x_2, \ldots, x_t)$ where each $x_i \in \mathcal{V}$.

\paragraph{Psyche state.} The psyche is a scalar $\psi_t \in [\psi_{\min}, \psi_{\max}] = [-100, +100]$, initialized at $\psi_0 = 0$. Negative values represent stress; positive values represent overconfidence; zero is the neutral (optimal) state.

\paragraph{Positional factors.} Five factor functions are defined $f_i : \mathcal{B} \to \mathbb{R}$ for $i \in \{1, \ldots, 5\}$, each evaluating the board from the perspective of the side to move (\Cref{tab:factors}).

\begin{table}[ht]
\centering
\small
\caption{Positional factor definitions for psyche computation.}
\label{tab:factors}
\begin{tabular}{@{}llll@{}}
\toprule
Factor & Symbol & Domain & Description \\
\midrule
Material & $f_1(b)$ & $\mathbb{Z}$ & $\sum_p v_p n_p^{\text{own}} - \sum_p v_p n_p^{\text{opp}}$, $\mathbf{v} = (1,3,3,5,9)$ \\
King safety & $f_2(b)$ & $\mathbb{Z}$ & Pawn shield minus enemy attackers in king's $3\!\times\!3$ area \\
Mobility & $f_3(b)$ & $\mathbb{Z}$ & $|\mathcal{L}^{\text{own}}| - |\mathcal{L}^{\text{opp}}|$ (own only if in check) \\
Center ctrl & $f_4(b)$ & $\mathbb{Z}$ & Own minus opp.\ attackers of $\{d4, d5, e4, e5\}$ \\
Vulnerability & $f_5(b)$ & $\mathbb{Z}_{\leq 0}$ & $-|\{s : \text{own piece on } s \text{ attacked}\}|$ \\
\bottomrule
\end{tabular}
\end{table}

\paragraph{Weight vector.} Each factor is scaled by a configurable weight $w_i \in \mathbb{R}_{\geq 0}$. Default values: $\mathbf{w} = (10, 5, 1, 2, 3)$.\footnote{All experiments in this paper use these values. The live Lichess deployment (\texttt{ailedbot}) uses updated tuning ($\mathbf{w} = (10, 3, 1, 2, 3)$, $s = 80$, $\gamma = 0.2$, $\phi_{\text{opening}} = 0.5$, $\phi_{\text{endgame}} = 0.8$) to produce wider psyche range in real games. Experiment parameters are frozen in the Zenodo archive.}

\paragraph{Check indicator.} The check indicator is $\chi(b) = -50$ if the side to move is in check, and $\chi(b) = 0$ otherwise.

\paragraph{Transformer notation.} I write $f_\theta$ for the move predictor (parameters $\theta$). The move predictor outputs logits $\mathbf{z}_t = f_\theta(\mathbf{x}_{1:t}) \in \mathbb{R}^{|\mathcal{V}|}$ over the full UCI move vocabulary $\mathcal{V}$.

% ============================================================================
% 4. THE PSYCHE MODEL
% ============================================================================
\section{The Psyche Model}\label{sec:psyche}

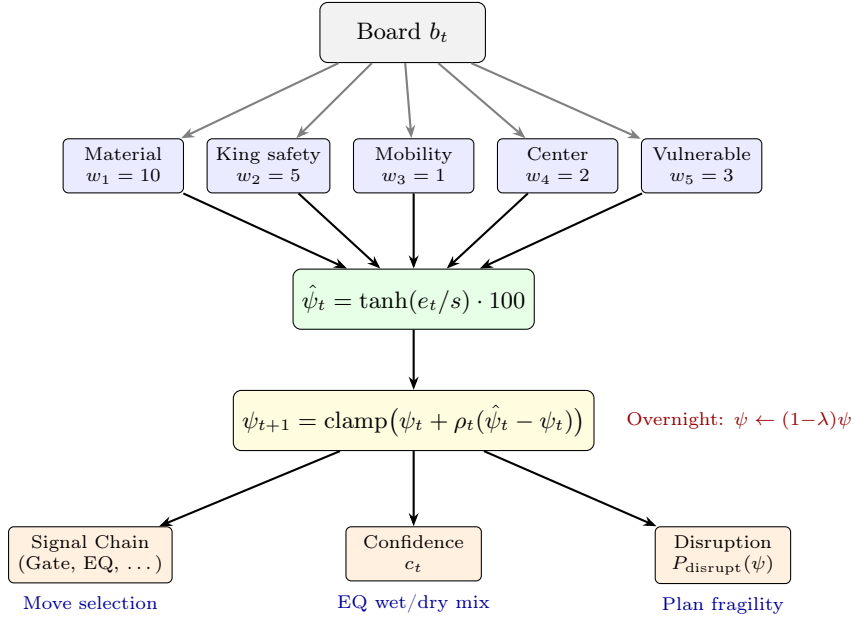
\begin{figure}[ht]
\centering
\begin{tikzpicture}[
    node distance=1.2cm and 1.5cm,
    box/.style={rectangle, draw, rounded corners=3pt, minimum width=2.2cm, minimum height=0.8cm, align=center, font=\small},
    factor/.style={rectangle, draw, fill=blue!8, rounded corners=2pt, minimum width=1.6cm, minimum height=0.6cm, align=center, font=\scriptsize},
    effect/.style={rectangle, draw, fill=orange!12, rounded corners=2pt, minimum width=1.8cm, minimum height=0.6cm, align=center, font=\scriptsize},
    arr/.style={-{Stealth[length=5pt]}, thick},
]

% Board position input
\node[box, fill=gray!10] (board) {Board $b_t$};

% Factors
\node[factor, below left=1.0cm and 1.8cm of board] (f1) {Material\\$w_1 = 10$};
\node[factor, right=0.3cm of f1] (f2) {King safety\\$w_2 = 5$};
\node[factor, right=0.3cm of f2] (f3) {Mobility\\$w_3 = 1$};
\node[factor, right=0.3cm of f3] (f4) {Center\\$w_4 = 2$};
\node[factor, right=0.3cm of f4] (f5) {Vulnerable\\$w_5 = 3$};

% Psyche delta
\node[box, fill=green!10, below=1.0cm of f3] (delta) {$\hat{\psi}_t = \tanh(e_t/s)\cdot 100$};

% Psyche state
\node[box, fill=yellow!15, below=0.8cm of delta] (psi) {$\psi_{t+1} = \clamp\bigl(\psi_t + \rho_t(\hat{\psi}_t - \psi_t)\bigr)$};

% Effects
\node[effect, below left=1.0cm and 0.8cm of psi] (temp) {Signal Chain\\(Gate, EQ, \ldots)};
\node[effect, below=1.0cm of psi] (conf) {Confidence\\$c_t$};
\node[effect, below right=1.0cm and 0.8cm of psi] (disrupt) {Disruption\\$P_{\text{disrupt}}(\psi)$};

% Arrows from board to factors
\foreach \f in {f1,f2,f3,f4,f5} {
    \draw[arr, gray] (board) -- (\f);
}

% Arrows from factors to delta
\foreach \f in {f1,f2,f3,f4,f5} {
    \draw[arr] (\f) -- (delta);
}

% Delta to psi
\draw[arr] (delta) -- (psi);

% Psi to effects
\draw[arr] (psi) -- (temp);
\draw[arr] (psi) -- (conf);
\draw[arr] (psi) -- (disrupt);

% Labels for effects
\node[font=\scriptsize, below=0.05cm of temp, text=blue!60!black] {Move selection};
\node[font=\scriptsize, below=0.05cm of conf, text=blue!60!black] {EQ wet/dry mix};
\node[font=\scriptsize, below=0.05cm of disrupt, text=blue!60!black] {Plan fragility};

% Decay annotation
\node[font=\scriptsize, right=0.3cm of psi, text=red!60!black] {Overnight: $\psi \leftarrow (1{-}\lambda)\psi$};

\end{tikzpicture}
\caption{Psyche computation pipeline. Five positional factors are extracted from the board, weighted, and summed to produce a raw evaluation $e_t$, which is tanh-compressed to a target psyche $\hat{\psi}_t \in [-100,+100]$.}
\label{fig:psyche-overview}
\end{figure}

At its core, the psyche system does one thing: it looks at the board, boils down what it sees into a single number, and uses that number to push and pull decision quality across the whole system. \Cref{fig:psyche-overview} lays out the pipeline.

\subsection{Psyche Update Dynamics}

The psyche state evolves after each move according to a weighted sum of positional factors, clamped to the domain $[\psi_{\min}, \psi_{\max}]$.

\begin{definition}[Psyche update]\label{def:psyche-update}
Given a board position $b_t \in \mathcal{B}$, weight vector $\mathbf{w} \in \mathbb{R}_{\geq 0}^5$, check indicator $\chi(b_t)$, scale $s > 0$, and reactivity $\rho \in [0,1]$, the psyche update blends the current state toward a position-derived target:
\begin{align}
    e_t &= \sum_{i=1}^{5} w_i \cdot f_i(b_t) + \chi(b_t) \label{eq:raw-eval} \\
    \hat{\psi}_t &= \tanh\!\left(\frac{e_t}{s}\right) \cdot 100 \label{eq:target-psi} \\
    \psi_{t+1} &= \clamp\!\left(\psi_t + \rho_t \cdot (\hat{\psi}_t - \psi_t),\; \psi_{\min},\; \psi_{\max}\right) \label{eq:psi-update} \\
    \rho_t &= \rho \cdot \phi_t \cdot (1 - \gamma) \label{eq:reactivity}
\end{align}
where $\hat{\psi}_t \in (-100, +100)$ is the position-implied target (the tanh output is an open interval; the clamp in \Cref{eq:psi-update} guarantees $\psi_{t+1} \in [-100, +100]$); $s = 50$ scales how much raw evaluation is needed to reach the extremes; $\rho_t$ is the effective reactivity scaling the blend rate by game phase $\phi_t \in \{0.15, 1.0, 0.1\}$ (opening, midgame, endgame) and resilience $\gamma = 0.4$. Default $\rho = 0.3$ gives $\rho_t \leq 0.18$ in midgame, bounding the per-move psyche change to $|\Delta\psi_t| \leq 36$.
\end{definition}

The clamp operation guarantees \textbf{boundedness}: $\psi_t \in [\psi_{\min}, \psi_{\max}]$ for all $t$, regardless of factor magnitudes (immediate from the definition of clamp).

\subsection{Temporal Decay}

The psyche runs on two clocks. During a session, it updates move-by-move through \Cref{def:psyche-update}. Between sessions---think of it as overnight rest---it decays multiplicatively back toward zero.

\begin{definition}[Daily decay]\label{def:decay}
Given decay rate $\lambda \in [0, 1]$, the inter-session update is:
\begin{equation}
    \psi^{\text{day}}_{k+1} = (1 - \lambda) \cdot \psi^{\text{day}}_k
\end{equation}
where $k$ indexes sessions and $\psi^{\text{day}}_k$ is the psyche at the end of session~$k$. \Cref{def:psyche-update} applies within each session; \Cref{def:decay} applies between sessions.
\end{definition}

Without intervening games, $\psi^{\text{day}}_k = (1 - \lambda)^k \cdot \psi^{\text{day}}_0 \to 0$, with half-life $k_{1/2} = \lceil \log(0.5) / \log(1 - \lambda) \rceil = 4$ days at $\lambda = 0.20$.

This decay is what gives the system its longer-term arc. A losing streak pushes $\psi$ negative, which raises the error rate, which causes more losses---a self-reinforcing spiral, not unlike tilt in poker or esports, held in check by the clamp at $\psi_{\min}$ and eventually unwound by overnight decay. The reverse happens too: a run of wins inflates $\psi$, the engine starts playing recklessly, and the resulting blunders can pull it back down.

\subsection{Signal Chain: Audio-Inspired Move Selection}\label{sec:signal-chain}

Here's the core analogy: shaping a probability distribution over chess moves isn't so different from running an audio signal through a chain of effects pedals. Each stage is a pure probability transform---no search, no game history---applied one after another to the raw model output. $\psi_t$ parameterizes every stage; a static personality preset pins down the character of each transform. The upshot is a clean separation. Personality says \emph{who} the engine is. Psyche says \emph{how} that character reacts to pressure.

The signal chain consists of eight stages (legal-move extraction plus seven probability transforms):

\begin{algorithm}[t]
\caption{Signal Chain: Psyche-Modulated Move Selection}\label{alg:move-selection}
\begin{algorithmic}[1]
\REQUIRE Full-vocabulary logits $\mathbf{z}_t \in \mathbb{R}^{|\mathcal{V}|}$, legal moves $\mathcal{L}(b_t)$, entropy confidence $c_H \in [0,1]$, psyche $\psi_t$, personality preset $\Pi$
\STATE $\mathbf{q} \leftarrow \mathbf{z}_t\bigl[\mathcal{L}(b_t)\bigr]$ \hfill \COMMENT{0. Legal-move extraction: index-select $|\mathcal{L}(b_t)|$ entries; all subsequent stages operate over this legal-move support only}
\STATE $\mathbf{p} \leftarrow \softmax(\mathbf{q})$ \hfill \COMMENT{1. Raw probabilities}
\STATE $\mathbf{p} \leftarrow \text{Gate}(\mathbf{p}, \tau_g(\psi_t, \Pi))$ \hfill \COMMENT{2. Noise gate}
\STATE $\mathbf{p} \leftarrow \text{Dynamics}(\mathbf{p}, \alpha(\psi_t, \Pi))$ \hfill \COMMENT{3. Compressor/expander}
\STATE $\mathbf{p} \leftarrow \text{EQ}(\mathbf{p}, \mathbf{g}(\psi_t, c_t, \Pi))$ \hfill \COMMENT{4. Equalizer bands}
\STATE $\mathbf{p} \leftarrow \text{Saturate}(\mathbf{p}, \sigma(\psi_t, \Pi))$ \hfill \COMMENT{5. Saturation ceiling}
\STATE $\mathbf{p} \leftarrow \mathbf{p} / \|\mathbf{p}\|_1$ \hfill \COMMENT{6. Renormalize}
\STATE $m^* \sim \text{Categorical}(\mathbf{p})$ \hfill \COMMENT{7. Sample}
\RETURN $m^*$
\end{algorithmic}
\end{algorithm}

The full signal chain is the composed function:
\begin{equation}
S(\mathbf{p}; \psi, \Pi) = \text{norm}\bigl(\text{sat}(\text{eq}(\text{dyn}(\text{gate}(\mathbf{p}; \psi, \Pi); \psi, \Pi); \psi, \Pi, c_H); \psi, \Pi)\bigr)
\end{equation}
where each stage is applied left-to-right and $c_H = 1 - H/H_{\max}$.

All parameters are defined at three psyche anchors (stress, neutral, overconfident) and linearly interpolated. For a parameter with anchor values $(v_s, v_n, v_c)$:
\begin{equation}\label{eq:lerp-psyche}
    v(\psi) = \begin{cases}
        v_n + (v_s - v_n) \cdot |\hat\psi| & \text{if } \hat\psi \leq 0 \\
        v_n + (v_c - v_n) \cdot \hat\psi & \text{if } \hat\psi > 0
    \end{cases}
    \quad\text{where}\quad \hat\psi = \clamp(\psi / 100, -1, 1)
\end{equation}

\subsubsection{Noise Gate}

Moves with probability below a psyche-dependent threshold $\tau_g(\psi)$ are silenced (zeroed):
\begin{equation}
    p_i' = \begin{cases} p_i & \text{if } p_i \geq \tau_g(\psi) \\ 0 & \text{otherwise} \end{cases}, \qquad
    \text{then} \quad \mathbf{p}' \leftarrow \mathbf{p}' / \|\mathbf{p}'\|_1
\end{equation}
If $\tau_g(\psi) > \max_i p_i$ (all moves would be silenced), the gate step is skipped and $\mathbf{p}' = \mathbf{p}$, preserving the full distribution. When $\psi$ is low (the stress end), the gate swings open and weak moves leak through. When $\psi$ is high (overconfidence), the gate tightens and only the stronger candidates get past. Defaults: $\tau_g = (0.005, 0.02, 0.06)$ for stress, neutral, and overconfident respectively.

At $\psi = 0$ (neutral), the threshold sits at $\tau_g^{(n)} = 0.02$---a moderate filter. It keeps most of the distribution intact but clips the long tail of moves nobody would seriously consider.

\subsubsection{Dynamics: Compressor/Expander}

A power-law transform reshapes the probability spread:
\begin{equation}\label{eq:dynamics}
    p_i' = \frac{p_i^{\alpha(\psi)}}{\sum_j p_j^{\alpha(\psi)}}
\end{equation}
where $\alpha(\psi)$ interpolates between stress and overconfident anchors with $\alpha_n = 1.0$ (unity at neutral). When $\alpha < 1$ the stage acts as an expander, flattening the distribution so that an engine at low $\psi$ (the stress end) considers a wider range of moves. When $\alpha > 1$ it compresses, sharpening the distribution so that an engine at high $\psi$ (the overconfidence end) concentrates on fewer candidates. Default anchors: $\alpha = (0.5, 1.0, 2.0)$.

\subsubsection{Equalizer Bands}

Legal moves get sorted by model probability and split into five rank bands---best, good, mild, bad, worst---each roughly the same size. Worth stressing: the bands come from the model's probability ranking, not from any objective evaluation of move quality. Each band receives a gain multiplier $g_k(\psi, c)$ that boosts or attenuates the moves within it:
\begin{equation}
    p_i' = p_i \cdot g_{k(i)}(\psi, c), \qquad k(i) = \text{band of move } i
\end{equation}
The gains are interpolated across three psyche states (stress, neutral, overconfident) per band, then blended with a flat reference using entropy confidence $c_H \in [0,1]$ as a wet/dry mix:
\begin{equation}\label{eq:eq-wetdry}
    g_k^{\text{eff}} = 1 + (g_k(\psi) - 1) \cdot c
\end{equation}
When $c = 0$ (the position is murky and the model isn't sure), EQ stays out of the way. At $c = 1$ (clear position, high confidence), EQ kicks in fully.

\Cref{tab:eq-bands} shows the default ``human'' personality preset. Under stress, the best moves are slightly suppressed while mid-quality moves are boosted---an effect loosely analogous to scattered attention. At neutral, top moves receive moderate boost. Overconfident play bypasses EQ entirely (flat gains), trusting raw model probabilities.

\begin{table}[ht]
\centering
\small
\caption{EQ band gains for the ``human'' personality preset across psyche states.}
\label{tab:eq-bands}
\begin{tabular}{@{}lccccc@{}}
\toprule
Psyche state & Best & Good & Mild & Bad & Worst \\
\midrule
Stress ($\psi = -100$) & 0.80 & 1.30 & 1.00 & 1.40 & 1.20 \\
Neutral ($\psi = 0$) & 1.30 & 1.20 & 1.00 & 0.70 & 0.50 \\
Overconfident ($\psi = +100$) & 1.00 & 1.00 & 1.00 & 1.00 & 1.00 \\
\bottomrule
\end{tabular}
\end{table}

The gains span a $4\times$ range ($g_k \in \{0.5, 1.0, 1.5, 2.0\}$), a range loosely informed by the scale of biases observed in human decision-making research \citep{kahneman1979prospect}. The five-band partition uses equal-count binning over legal moves (not equal-probability binning) to ensure each band always has at least one representative move.

\subsubsection{Saturation}

Finally, a probability ceiling $\sigma(\psi)$ keeps any one move from running away with all the probability mass:
\begin{equation}
    p_i' = \min(p_i, \sigma(\psi))
\end{equation}
followed by renormalization. At the stress end, the ceiling is low---no single move gets to dominate. Under overconfidence, the ceiling rises and near-deterministic play becomes possible. Defaults: $\sigma = (0.30, 0.50, 0.85)$.

\subsubsection{Personality Presets}

Each personality preset $\Pi$ bundles together the full parameter set: EQ gains at three psyche states, dynamics power anchors, gate thresholds, saturation ceilings. Six personality presets are defined inspired by music genres (\Cref{tab:presets}): \emph{flat} (bypass), \emph{classical} (disciplined, tight gate), \emph{rock} (bold, V-shape EQ), \emph{jazz} (avoids the obvious, heavy saturation), \emph{metal} (chaotic, boosts worst moves), and \emph{human} (default, stress-sensitive).

\begin{table}[ht]
\centering
\small
\caption{Personality presets: dynamics and gate/saturation anchors (stress / neutral / overconfident).}
\label{tab:presets}
\begin{tabular}{@{}lcccc@{}}
\toprule
Preset & Dynamics $\alpha$ & Gate $\tau_g$ & Saturation $\sigma$ & Character \\
\midrule
flat & 1.0 / 1.0 & 0 / 0 / 0 & 1.0 / 1.0 / 1.0 & Bypass \\
classical & 0.8 / 1.5 & .01 / .02 / .05 & .50 / .60 / .85 & Disciplined \\
rock & 0.6 / 1.6 & .005 / .01 / .03 & .35 / .45 / .70 & Bold \\
jazz & 0.5 / 1.4 & .001 / .005 / .01 & .25 / .35 / .50 & Creative \\
metal & 0.4 / 1.3 & 0 / .001 / .005 & .20 / .30 / .50 & Chaotic \\
human & 0.5 / 2.0 & .005 / .02 / .06 & .30 / .50 / .85 & Realistic \\
\bottomrule
\end{tabular}
\end{table}

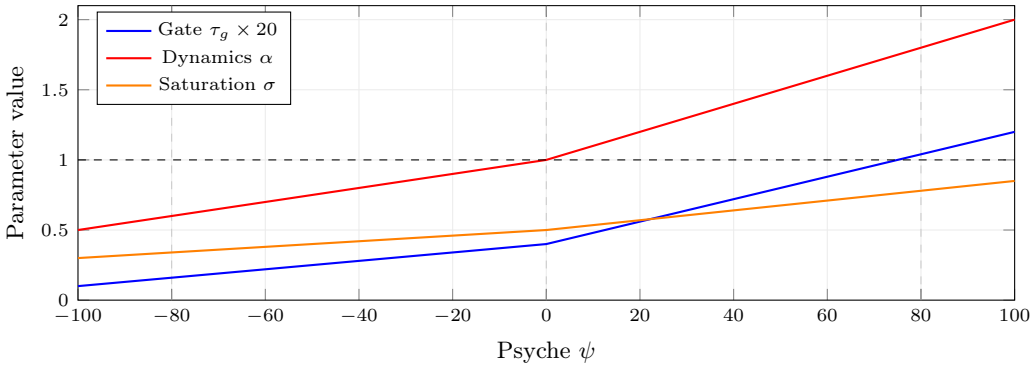
\begin{figure}[ht]
\centering
\begin{tikzpicture}
\begin{axis}[
    width=0.85\textwidth,
    height=5.5cm,
    xlabel={Psyche $\psi$},
    ylabel={Parameter value},
    xmin=-100, xmax=100,
    ymin=0, ymax=2.1,
    legend style={font=\scriptsize, at={(0.02,0.98)}, anchor=north west},
    grid=major,
    grid style={gray!15},
    tick label style={font=\scriptsize},
    label style={font=\small},
    extra x ticks={-80,0,80},
    extra x tick style={grid=major, grid style={dashed, gray!40}},
    extra x tick labels={},
]
% Gate threshold (scaled ×20 for visibility)
\addplot[thick, blue, mark=none] coordinates {(-100,0.10) (-90,0.13) (-80,0.16) (-70,0.19) (-60,0.22) (-50,0.25) (-40,0.28) (-30,0.31) (-20,0.34) (-10,0.37) (0,0.40) (10,0.48) (20,0.56) (30,0.64) (40,0.72) (50,0.80) (60,0.88) (70,0.96) (80,1.04) (90,1.12) (100,1.20)};
\addlegendentry{Gate $\tau_g \times 20$}
% Dynamics power
\addplot[thick, red, mark=none] coordinates {(-100,0.50) (-90,0.55) (-80,0.60) (-70,0.65) (-60,0.70) (-50,0.75) (-40,0.80) (-30,0.85) (-20,0.90) (-10,0.95) (0,1.00) (10,1.10) (20,1.20) (30,1.30) (40,1.40) (50,1.50) (60,1.60) (70,1.70) (80,1.80) (90,1.90) (100,2.00)};
\addlegendentry{Dynamics $\alpha$}
% Saturation ceiling
\addplot[thick, orange, mark=none] coordinates {(-100,0.300) (-90,0.320) (-80,0.340) (-70,0.360) (-60,0.380) (-50,0.400) (-40,0.420) (-30,0.440) (-20,0.460) (-10,0.480) (0,0.500) (10,0.535) (20,0.570) (30,0.605) (40,0.640) (50,0.675) (60,0.710) (70,0.745) (80,0.780) (90,0.815) (100,0.850)};
\addlegendentry{Saturation $\sigma$}
% Unity reference line
\addplot[thin, dashed, black] coordinates {(-100,1.0) (100,1.0)};
\end{axis}
\end{tikzpicture}
\caption{Signal chain parameters for the ``human'' personality preset as a function of psyche $\psi$. Gate threshold is scaled $\times 20$ for visibility. The dynamics power $\alpha$ crosses unity at $\psi = 0$: below neutral it acts as an expander (flattening the distribution), above as a compressor (sharpening it). All parameters interpolate linearly between the three psyche anchors.}
\label{fig:signal-params}
\end{figure}

\Cref{fig:signal-params} shows how the three scalar parameters of the signal chain (gate threshold, dynamics power, saturation ceiling) evolve as psyche moves from stress ($-100$) to overconfidence ($+100$) for the ``human'' preset. All three interpolate linearly across the three psyche anchors (\Cref{eq:lerp-psyche}). The dynamics power crosses unity at $\psi = 0$: below neutral the distribution flattens (expander), above it sharpens (compressor).

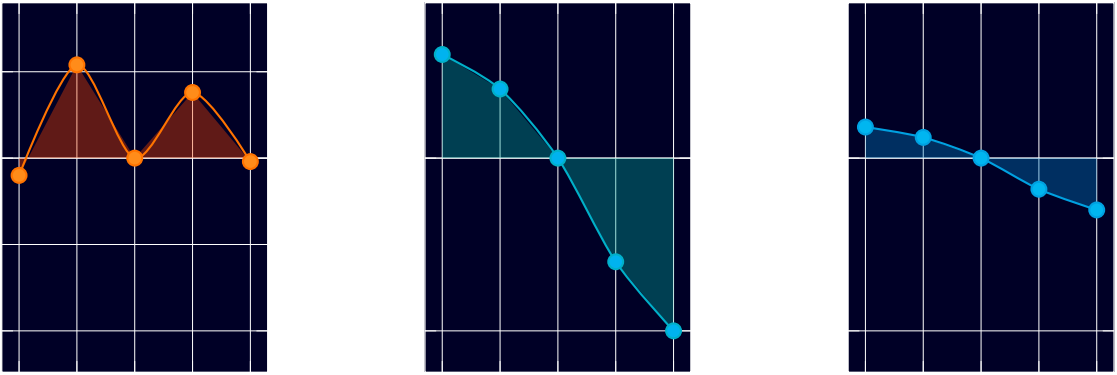
\begin{figure}[ht]
\centering
\pgfplotsset{
  eqpanel/.style={
    width=0.31\textwidth,
    height=6.5cm,
    xmin=-0.3, xmax=4.3,
    ymin=-0.62, ymax=0.45,
    xtick={0,1,2,3,4},
    xticklabels={Best,Good,Mild,Bad,Worst},
    x tick label style={font=\scriptsize, color=white!75},
    ytick={-0.50,-0.25,0,0.25},
    yticklabel style={font=\scriptsize, color=white!55},
    title style={font=\small\bfseries, color=white!92},
    ymajorgrids=true,
    xmajorgrids=true,
    grid style={white!10, line width=0.25pt},
    axis background/.style={fill=black!85!blue},
    axis line style={color=white!18, line width=0.5pt},
    tick style={color=white!18},
  }
}
%
% Panel 1 — Stress (psi=-70): humps at Good and Bad
% Best=-0.05  Good=+0.27  Mild=0  Bad=+0.19  Worst=-0.01
\begin{tikzpicture}
\begin{axis}[eqpanel,
  title={Stress ($\psi=-70$)},
  ylabel={Gain$\,-\,$1},
  ylabel style={font=\scriptsize, color=white!60},
]
\addplot[thin, white!50, forget plot] coordinates {(-0.3,0) (4.3,0)};
\addplot[draw=none, fill=orange!55!red, fill opacity=0.38]
    coordinates {(0,0)(0,-0.05)(1,0.27)(2,0.00)(3,0.19)(4,-0.01)(4,0)} \closedcycle;
\addplot[smooth, thick, orange!90!red, forget plot,
         mark=*, mark size=2.8pt, mark options={fill=orange!90!white, draw=none}]
    coordinates {(0,-0.05) (1,0.27) (2,0.00) (3,0.19) (4,-0.01)};
\end{axis}
\end{tikzpicture}%
\hfill%
%
% Panel 2 — Neutral (psi=0): monotone preference ordering
% Best=+0.30  Good=+0.20  Mild=0  Bad=-0.30  Worst=-0.50
\begin{tikzpicture}
\begin{axis}[eqpanel,
  title={Neutral ($\psi=0$)},
  ytick={},
]
\addplot[thin, white!50, forget plot] coordinates {(-0.3,0) (4.3,0)};
\addplot[draw=none, fill=cyan!50!green, fill opacity=0.38]
    coordinates {(0,0)(0,0.30)(1,0.20)(2,0.00)(3,-0.30)(4,-0.50)(4,0)} \closedcycle;
\addplot[smooth, thick, cyan!75!green!90, forget plot,
         mark=*, mark size=2.8pt, mark options={fill=cyan!90!white, draw=none}]
    coordinates {(0,0.30) (1,0.20) (2,0.00) (3,-0.30) (4,-0.50)};
\end{axis}
\end{tikzpicture}%
\hfill%
%
% Panel 3 — Overconfident (psi=+70): nearly flat
% Best=+0.09  Good=+0.06  Mild=0  Bad=-0.09  Worst=-0.15
\begin{tikzpicture}
\begin{axis}[eqpanel,
  title={Overconfident ($\psi=+70$)},
  ytick={},
]
\addplot[thin, white!50, forget plot] coordinates {(-0.3,0) (4.3,0)};
\addplot[draw=none, fill=cyan!60!blue, fill opacity=0.38]
    coordinates {(0,0)(0,0.09)(1,0.06)(2,0.00)(3,-0.09)(4,-0.15)(4,0)} \closedcycle;
\addplot[smooth, thick, cyan!85!blue!90, forget plot,
         mark=*, mark size=2.8pt, mark options={fill=cyan!85!white, draw=none}]
    coordinates {(0,0.09) (1,0.06) (2,0.00) (3,-0.09) (4,-0.15)};
\end{axis}
\end{tikzpicture}%
\caption{EQ band gains for the ``human'' personality preset at three psyche conditions, displayed as equalizer panels. Each dot is a band control point (Best, Good, Mild, Bad, Worst); the smooth interpolated curve and shaded fill show the gain deviation from the white unity baseline (above\,$=$\,boost; below\,$=$\,cut). Under stress ($\psi=-70$), humps at ``good'' and ``bad'' shift weight toward adjacent tiers. At neutral ($\psi=0$), a strong monotone preference ordering is imposed. At overconfidence ($\psi=+70$), all deviations converge toward zero and the EQ becomes nearly flat.}
\label{fig:eq-evolution}
\end{figure}

\Cref{fig:eq-evolution} renders the EQ gain profile for each psyche condition in a separate panel styled as a hardware equalizer display. Each dot marks a band control point; the smooth interpolated curve and its shaded fill show the gain deviation from the white unity baseline (above $=$ boost; below $=$ cut). The \emph{Mild} band sits at zero deviation in all three conditions.

\paragraph{Engine-agnostic design.} The chain only ever sees a probability distribution over legal moves. It doesn't need a search tree, doesn't look at game history, and has no engine-specific hooks. If your engine outputs move probabilities, you can wrap it in this chain. I demonstrate exactly that with Maia2 in \Cref{sec:experiments}.

\paragraph{Implementation details.}
Legal-move extraction occurs before softmax: the predictor outputs logits over $\mathcal{V}$, from which Ailed index-selects the entries corresponding to legal moves. All subsequent transforms operate over this legal-move support. The noise gate includes a safety check: if the threshold would silence all moves, the gate step is skipped and the full distribution is preserved.

The psyche state $\psi$ is updated after every ply (both sides) via the five-factor calculator (\Cref{sec:psyche}). The final update before move selection is always from Ailed's perspective.

Entropy confidence $c_H = 1 - H/H_{\max}$ (where $H$ is Shannon entropy in nats over legal moves and $H_{\max} = \ln N$) is the sole confidence signal for the EQ wet/dry mix at inference time.

% ============================================================================
% 5. ARCHITECTURE
% ============================================================================
\section{Architecture}\label{sec:architecture}

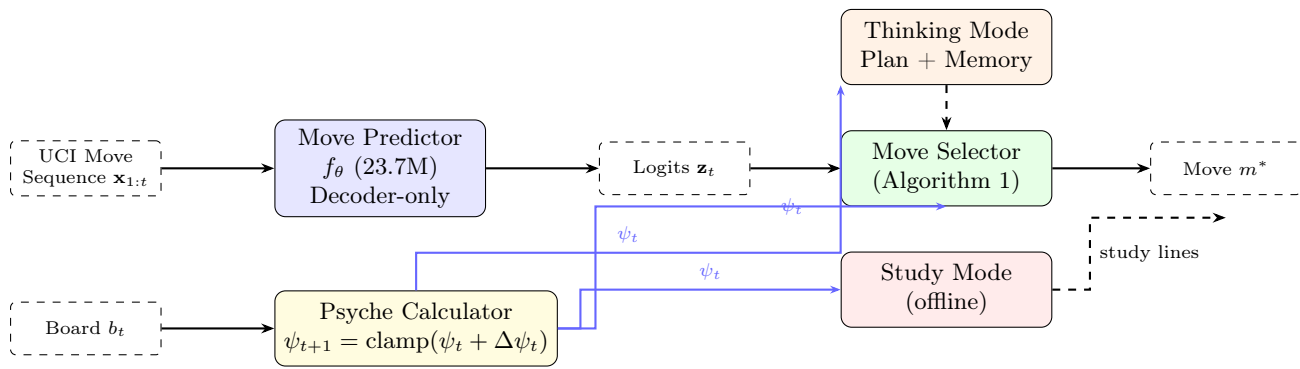
\begin{figure}[ht]
\centering
\begin{tikzpicture}[
    node distance=0.9cm and 1.2cm,
    module/.style={rectangle, draw, rounded corners=4pt, minimum width=2.8cm, minimum height=1.0cm, align=center, font=\small},
    io/.style={rectangle, draw, dashed, rounded corners=2pt, minimum width=2.0cm, minimum height=0.7cm, align=center, font=\scriptsize},
    arr/.style={-{Stealth[length=5pt]}, thick},
    dataarr/.style={-{Stealth[length=4pt]}, thick, blue!60},
]

% Input
\node[io] (input) {UCI Move\\Sequence $\mathbf{x}_{1:t}$};

% Move predictor (sole neural module)
\node[module, fill=blue!10, right=1.5cm of input] (mp) {Move Predictor\\$f_\theta$ (23.7M)\\Decoder-only};

% Board state (independent input to psyche)
\node[io, below=1.4cm of input] (board) {Board $b_t$};

% Psyche calculator
\node[module, fill=yellow!15, right=1.5cm of board] (psyche) {Psyche Calculator\\$\psi_{t+1} = \clamp(\psi_t + \Delta\psi_t)$};

% Logits
\node[io, right=1.5cm of mp] (logits) {Logits $\mathbf{z}_t$};

% Move selector
\node[module, fill=green!10, right=1.2cm of logits] (selector) {Move Selector\\(\Cref{alg:move-selection})};

% Cognitive modules
\node[module, fill=orange!10, above=0.6cm of selector] (thinking) {Thinking Mode\\Plan + Memory};
\node[module, fill=red!8, below=0.6cm of selector] (study) {Study Mode\\(offline)};

% Final output
\node[io, right=1.3cm of selector] (output) {Move $m^*$};

% Arrows
\draw[arr] (input) -- (mp);
\draw[arr] (board) -- (psyche);
\draw[arr] (mp) -- (logits);
\draw[arr] (logits) -- (selector);
\draw[dataarr] (psyche.east) -- ++(0.5,0) |- node[font=\scriptsize, near end, right] {$\psi_t$} (selector.south);
\draw[dataarr] (psyche.north) -- ++(0, 0.5) -| node[font=\scriptsize, near start, above] {$\psi_t$} (thinking.south west);
\draw[dataarr] (psyche.east) -- ++(0.3, 0) |- node[font=\scriptsize, near end, above] {$\psi_t$} (study.west);
\draw[arr] (selector) -- (output);
\draw[arr, dashed] (thinking) -- (selector);
\draw[arr, dashed] (study.east) -- ++(0.5, 0) |- node[font=\scriptsize, near start, right] {study lines} ([yshift=-0.3cm]output.south);

\end{tikzpicture}
\caption{Ailed system architecture. The move predictor $f_\theta$ produces logits fed into the move selector; the psyche calculator computes the psyche state $\psi_t$ from board position evaluations. Psyche state modulates all three output paths: move selection (signal chain), thinking mode (plan disruption), and study mode (skip probability and exploration quality). Dashed arrows indicate conditional paths.}
\label{fig:architecture}
\end{figure}

In principle, the psyche model can wrap any system that spits out move logits. For this paper I pair it with a single transformer for move prediction, and derive entropy confidence straight from the move distribution. \Cref{fig:architecture} shows how everything fits together.

\subsection{Move Predictor}

The move predictor $f_\theta$ is a standard decoder-only transformer \citep{vaswani2017attention}. It models $p(x_{t+1} \mid \mathbf{x}_{1:t})$---the conditional distribution over the next move, given a history of UCI move tokens.

Input tokens $\mathbf{x}_{1:t}$ are embedded via a learned embedding matrix $\mathbf{E} \in \mathbb{R}^{|\mathcal{V}| \times d}$ and scaled by $\sqrt{d}$. Sinusoidal positional encodings $\mathbf{PE} \in \mathbb{R}^{L_{\max} \times d}$ are added:
\begin{equation}
    \mathbf{h}_0 = \text{Dropout}\!\left(\mathbf{E}[\mathbf{x}_{1:t}] \cdot \sqrt{d} + \mathbf{PE}_{1:t}\right)
\end{equation}

The sequence passes through $N$ transformer layers, each consisting of masked multi-head self-attention and a position-wise feedforward network with residual connections and layer normalization:
\begin{equation}
    \mathbf{h}_\ell = \text{LN}\!\left(\mathbf{h}_{\ell-1} + \text{FFN}\!\left(\text{LN}\!\left(\mathbf{h}_{\ell-1} + \text{MHA}(\mathbf{h}_{\ell-1})\right)\right)\right)
\end{equation}

A causal mask $\mathbf{M} \in \{0, -\infty\}^{t \times t}$ with $M_{ij} = -\infty$ for $j > i$ enforces autoregressive prediction. The output logits are:
\begin{equation}
    \mathbf{z}_t = \mathbf{W}_{\text{out}} \, \mathbf{h}_N[t] + \mathbf{b}_{\text{out}}, \quad \mathbf{z}_t \in \mathbb{R}^{|\mathcal{V}|}
\end{equation}

\begin{table}[ht]
\centering
\small
\caption{Move predictor hyperparameters.}
\label{tab:move-predictor}
\begin{tabular}{@{}ll@{}}
\toprule
Hyperparameter & Value \\
\midrule
Vocabulary $|\mathcal{V}|$ & 4,547 \\
Embedding dim $d$ & 512 \\
Attention heads & 8 \\
Layers $N$ & 6 \\
FFN dim & 2,048 \\
Max sequence length $L_{\max}$ & 300 \\
Parameters $|\theta|$ & 23.6M \\
\bottomrule
\end{tabular}
\end{table}

This one model does triple duty: it generates moves during play, produces multi-step plans in thinking mode (\Cref{sec:thinking}), and explores alternative lines in study mode (\Cref{sec:study}).

\subsection{Training}\label{sec:training}

Training data comes from the Lichess open database \citep{lichess_database}, filtered down to the \textbf{1025--1175 ELO} band. I pull roughly 60{,}000 games from standard-rated monthly archives, applying symmetric difficulty sampling and side-balanced selection, then split them deterministically: ${\sim}$48{,}000 for training, 5{,}000 for validation, 5{,}000 for testing.

The move predictor trains on next-token prediction with cross-entropy loss:
\begin{equation}
    \mathcal{L}_{\text{move}} = -\frac{1}{T}\sum_{t=1}^{T} \log p_\theta(x_t \mid \mathbf{x}_{1:t-1})
\end{equation}
Training goes for 44 epochs with early stopping (patience~5). Best validation loss: 3.80; validation accuracy: 40.7\%. I use AdamW ($\text{lr} = 10^{-4}$, weight decay $0.01$), batch size 32, gradient clipping at 1.0, Xavier initialization. The whole thing trains in about 32 minutes on an Apple M-series GPU.

\textbf{Confidence signal.} The only confidence signal in the system is entropy confidence, $c_H = 1 - H/H_{\max}$ (\Cref{sec:signal-chain}). I originally had an outcome predictor providing a second signal, but Experiment~C (\Cref{sec:ablation}) showed it wasn't pulling its weight. It's been removed.

\paragraph{Progressive learning.} Once the initial training is done, Ailed keeps learning. It plays games, accumulates experience, and every week (provided at least 20~games have been played) the move predictor gets fine-tuned on a mix of recent games and study data (\Cref{sec:study}), with the learning rate dropped to $1 \times 10^{-5}$. Play $\to$ study $\to$ retrain $\to$ play.

% ============================================================================
% 6. COGNITIVE EXTENSIONS
% ============================================================================
\section{Cognitive Extensions}\label{sec:cognitive}

So far the psyche model has only touched move selection. But the whole point of having a scalar that tracks ``how things are going'' is that you can pipe it into other parts of the system too. I do that here with two higher-order processes: lookahead planning and offline study. Neither is experimentally validated yet---I include them as design descriptions. Evaluation through live Lichess play is underway and will be covered in a follow-up.

\subsection{Thinking Mode: Fragile Lookahead Planning}\label{sec:thinking}

\begin{figure}[ht]
\centering
\begin{tikzpicture}[
    node distance=0.7cm and 1.0cm,
    state/.style={rectangle, draw, rounded corners=3pt, minimum width=2.0cm, minimum height=0.7cm, align=center, font=\scriptsize},
    decision/.style={diamond, draw, aspect=2.2, minimum width=1.2cm, align=center, font=\scriptsize},
    arr/.style={-{Stealth[length=4pt]}, thick},
]

\node[state, fill=blue!10] (start) {Confidence\\$c_t > 0.70$?};
\node[state, fill=green!10, right=1.2cm of start] (gen) {Generate plan\\($2D{-}1$ moves)};
\node[state, fill=yellow!10, right=1.2cm of gen] (store) {Store in\\memory $\mathcal{B}$};
\node[state, fill=orange!10, below=0.8cm of store] (check) {Opponent move\\matches?};
\node[state, fill=red!10, left=1.2cm of check] (discard) {Discard plan\\$\to$ normal};
\node[state, fill=purple!10, right=1.2cm of check] (roll) {Roll disruption\\$P_{\text{disrupt}}(\psi)$};
\node[state, fill=green!15, below=0.8cm of roll] (exec) {Execute\\planned move};

\draw[arr] (start) -- node[font=\scriptsize, above] {yes} (gen);
\draw[arr] (start.south) -- ++(0, -0.4) -| node[font=\scriptsize, near start, right] {no} (discard);
\draw[arr] (gen) -- (store);
\draw[arr] (store) -- node[font=\scriptsize, right] {next turn} (check);
\draw[arr] (check) -- node[font=\scriptsize, above] {no} (discard);
\draw[arr] (check) -- node[font=\scriptsize, above] {yes} (roll);
\draw[arr] (roll) -- node[font=\scriptsize, right] {pass} (exec);
\draw[arr] (roll.west) -- ++(-0.5, 0) |- node[font=\scriptsize, near end, above] {disrupted} (discard);

\end{tikzpicture}
\caption{Thinking mode lifecycle. A plan is generated when confidence exceeds the threshold, stored in a single-slot buffer, and executed only if the opponent's actual move matches the prediction \emph{and} the psyche disruption check passes.}
\label{fig:thinking}
\end{figure}
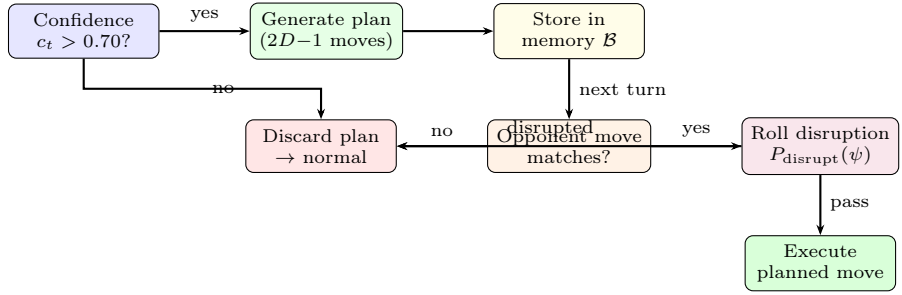

When a human chess player thinks ahead, it's usually a short linear sequence---``I go here, they go there, I take.'' Not a search tree. And the plan is fragile: one unexpected opponent move and it's gone, or sometimes the player is just too rattled to keep the thread. Thinking mode takes its cues from this. \Cref{fig:thinking} shows the lifecycle.

\paragraph{Plan generation.} When the entropy confidence $c_H$ (defined in \Cref{sec:signal-chain}) exceeds a threshold $c_{\text{plan}} = 0.70$, the move predictor generates a plan autoregressively. Given lookahead depth~$D$ (default $D = 2$), the planner produces $D$ own moves interleaved with $D - 1$ predicted opponent responses, totaling $2D - 1$ moves. Each predicted move is validated against a simulated board; if any move is illegal, plan generation aborts.

Plans live in a single-slot buffer $\mathcal{B}_{\text{plan}}$. Making a new plan wipes the old one---you only get one line of thought at a time.

\paragraph{Plan verification.} On subsequent turns, the engine compares the opponent's actual move $m_{\text{actual}}$ to the predicted move $m_{\text{predicted}}$ stored in $\mathcal{B}_{\text{plan}}$. If $m_{\text{actual}} \neq m_{\text{predicted}}$, the plan is discarded and normal move selection resumes. If $m_{\text{actual}} = m_{\text{predicted}}$, a psyche disruption check is rolled before executing the planned response.

\paragraph{Psyche-based disruption.} Even when predictions are correct, psyche extremes can disrupt plan execution.

\begin{definition}[Plan disruption]\label{def:disruption}
Given psyche $\psi_t$, disruption threshold $\tau_d \geq 0$, and pole-dependent maximum rates $r^{-}, r^{+} \in [0,1]$:
\begin{equation}
    P_{\text{disrupt}}(\psi_t) = \begin{cases}
        0 & \text{if } |\psi_t| \leq \tau_d \\[6pt]
        \dfrac{|\psi_t| - \tau_d}{100 - \tau_d} \cdot r^{-} & \text{if } \psi_t < -\tau_d \\[6pt]
        \dfrac{|\psi_t| - \tau_d}{100 - \tau_d} \cdot r^{+} & \text{if } \psi_t > \tau_d
    \end{cases}
\end{equation}
With defaults $\tau_d = 0$, $r^{-} = 0.80$, $r^{+} = 0.60$.
\end{definition}

The asymmetry is deliberate. Stress disrupts planning harder than overconfidence ($r^{-} > r^{+}$), which is motivated by evidence that anxiety tends to impair working memory more than confidence does \citep{eysenck2007anxiety}. Concretely: at $\psi = -100$ there's an 80\% chance the plan gets ``forgotten'' under pressure; at $\psi = +100$ there's a 60\% chance it gets ``ignored'' in favor of playing on impulse.

\subsection{Study Mode: Psyche-Degraded Offline Learning}\label{sec:study}

Strong players get better by reviewing their losses. Study mode captures this: after a loss, Ailed replays the game, finds the critical turning points, and explores what it could have done differently. The catch---and this is the important design choice---is that study quality degrades at psyche extremes. If the engine is tilted, it studies badly.

\paragraph{Turning point detection.} The detector replays a lost game move by move, computing the psyche trajectory $\psi_0, \psi_1, \ldots, \psi_N$. Turning points are the $K$ positions (default $K = 5$) with the largest negative psyche deltas:
\begin{equation}
    \mathcal{T} = \argtopK_{t}\left(-\Delta\psi_t\right), \quad \Delta\psi_t = \psi_{t+1} - \psi_t
\end{equation}

What this finds are positions that \emph{felt} bad---places where the psyche value dropped sharply---which isn't the same thing as positions that are objectively worst by engine evaluation. It's a bit like how some players remember their mistakes: not from careful post-game analysis, but because of how much the moment hurt.

\paragraph{Line exploration.} At each turning point $t \in \mathcal{T}$, the explorer: (1)~extracts the top-$K_a$ alternative first moves from the predictor logits (default $K_a = 5$); (2)~for each alternative, plays out $D_e$ moves autoregressively (default $D_e = 10$) through the signal chain at the current $\psi$; (3)~evaluates the final position using the signal chain's entropy confidence $c_H$; (4)~retains lines where the terminal $c_H$ of the alternative exceeds that of the original line.

The exploration uses the same psyche-modulated signal chain as game play (\Cref{sec:signal-chain}), creating a unified mechanism: at low $\psi$ Ailed explores too randomly (open gate, flat dynamics); at high $\psi$ it explores too narrowly (tight gate, compressed dynamics).

\paragraph{Psyche effects on study sessions.} Three channels degrade study quality at psyche extremes:

\begin{definition}[Study skip probability]\label{def:skip}
Given skip threshold $\tau_s$ and maximum skip rate $r_s$:
\begin{equation}
    P_{\text{skip}}(\psi) = \begin{cases}
        0 & \text{if } |\psi| \leq \tau_s \\[6pt]
        \dfrac{|\psi| - \tau_s}{100 - \tau_s} \cdot r_s & \text{otherwise}
    \end{cases}
\end{equation}
With defaults $\tau_s = 60$, $r_s = 0.70$. The threshold $\tau_s = 60$ permits unrestricted study within the central $|\psi| \leq 60$ range (mild stress or moderate overconfidence); the maximum rate $r_s = 0.70$ ensures study still occurs with 30\% probability even at extreme psyche, preventing complete learning shutdown.
\end{definition}

\begin{definition}[Effective study duration]\label{def:duration}
Given base timeout $D_{\text{base}}$ minutes:
\begin{equation}
    D_{\text{eff}}(\psi) = D_{\text{base}} \cdot \left(1 - \frac{|\psi|}{200}\right)
\end{equation}
At neutral psyche: full duration ($D_{\text{base}} = 60$~min). At extremes: half duration (30~min).
\end{definition}

\paragraph{Example.} Ailed at $\psi = -80$ after a losing session: $P_{\text{skip}} = \frac{80 - 60}{100 - 60} \cdot 0.70 = 0.35$ (35\% chance of skipping). If study proceeds: $D_{\text{eff}} = 60 \cdot (1 - 80/200) = 36$~minutes. Exploration runs through the signal chain at $\psi = -80$: the dynamics expander ($\alpha = 0.6$) flattens the distribution and the open gate ($\tau_g = 0.008$) admits weak moves, producing noisy alternatives.

\subsection{Weekly Fine-Tuning Loop}\label{sec:finetuning}

Study data integrates into the training pipeline through a configurable mixing ratio:
\begin{equation}
    \mathcal{D}_{\text{train}} = (1 - \rho) \cdot \mathcal{D}_{\text{games}} + \rho \cdot \mathcal{D}_{\text{study}}
\end{equation}
with default $\rho = 0.30$. If no study data exists, training uses $\rho = 0$. Fine-tuning runs weekly (minimum 20~games) with a reduced learning rate of $1 \times 10^{-5}$, creating the closed loop: play $\to$ study $\to$ retrain $\to$ play.

% ============================================================================
% 7. EXPERIMENTS
% ============================================================================
\section{Experiments}\label{sec:experiments}

The evaluation spans four experiments and 12{,}414~games, all played against vanilla Maia2-1100 \citep{tang2024maia2}. Experiments~A and~B (4{,}002~games combined) test three forced psyche conditions with two different move predictors. Experiment~C (2{,}400~games) ablates the confidence signal. Experiment~D (6{,}012~games) pulls the signal chain apart stage by stage.

The key design idea is that Experiments~A and~B run the \emph{exact same} psyche system and signal chain---only the probability source underneath changes. Ailed-60k (23.7M parameters, ${\sim}$60K training games) is the minimal-data case; Maia2+Psyche (23.3M parameters, 169M games) is the strong-model case. If the signal chain really works as engine-agnostic middleware, both should produce the same qualitative gradient.

% ----------------------------------------------------------------------------
% EXPERIMENTAL DESIGN
% ----------------------------------------------------------------------------
\subsection{Experimental Design}\label{sec:exp-design}

\paragraph{Match protocol.} Each experiment runs 2{,}001~games: three psyche conditions, 667~games each, colors balanced. Maia2 isn't a UCI engine---it's a Python library that hands back move probability distributions for a given board and ELO. I wrote a custom match runner to wrap both players. Vanilla Maia2-1100 always picks its top-probability move. Games end on checkmate, stalemate, threefold repetition, the fifty-move rule, insufficient material, or 400~plies.

\paragraph{Experiment~A: Ailed-60k.} On the psyche-modulated side: Ailed's own move predictor (23.7M parameters, trained on ${\sim}$60K games, 1025--1175 ELO). Signal chain running the ``human'' personality preset, entropy-based confidence.

\paragraph{Experiment~B: Maia2+Psyche.} Here the psyche-modulated side uses Maia2's own move probabilities. I convert them to log-probabilities ($z_i = \log p_i$) and pass those as logits to Algorithm~\ref{alg:move-selection}. Since $\mathrm{softmax}(\log \mathbf{p})_i = p_i$, I recover the original distribution exactly before any signal-chain processing touches it. The point of this setup is to test the chain as a pure personality layer---Maia2 brings strong predictions, and the psyche system controls how they get sampled.

\paragraph{Psyche initialization.} Three conditions: stress ($\psi_0 = -80$), neutral ($\psi_0 = 0$), and overconfident ($\psi_0 = +80$). After the first move, psyche evolves naturally through the position-based update rule (\Cref{sec:psyche}).

\paragraph{Per-move metrics.} At each modulated-side move, I record psyche~$\psi$, entropy confidence, Shannon entropy, and the engine's top-probability move before signal chain processing (for agreement analysis).

\paragraph{Implementation details.}
Legal moves are extracted by index-selection before softmax (Step~0 of Algorithm~\ref{alg:move-selection}); all subsequent stages operate over this legal-move support exclusively.
The gate step is skipped when its threshold would silence all legal moves.
Psyche is updated after every ply (both sides' moves) using the side-to-move perspective, so the value entering move selection always reflects Ailed's own view.
Confidence is entropy-based: $c_H = 1 - H/H_{\max}$ where
$H = -\sum_m p(m)\ln p(m)$ (Shannon entropy in nats over legal moves, with convention $0\ln 0 = 0$) and
$H_{\max} = \ln N$ is the maximum Shannon entropy achieved by the uniform distribution over $N$ legal moves ($p_i = 1/N$ for all $i$: $H_{\text{unif}} = -N\cdot\tfrac{1}{N}\ln\tfrac{1}{N} = \ln N$).

\paragraph{Human CPL baseline.}
To contextualize Ailed's per-move error rates, a human baseline was computed from Lichess standard-rated games (January--May 2016) filtered to the 950--1250~ELO band. After quality filters (rated games only, minimum 10 moves, no forfeits), 7,748 games were retained. Each move was evaluated at Stockfish depth~8, and psyche zones were assigned using the same factor-based boundaries as the engine experiments. Up to 25,000 moves per zone were sampled to equalize representation. Per-color psyche tracking---each side's psyche is updated independently from its own perspective---ensures that the zone assignment reflects the decision-maker's board state rather than the global evaluation. Full methodology is described in \Cref{sec:cpl-layer}.

% ----------------------------------------------------------------------------
% RESULTS
% ----------------------------------------------------------------------------
\subsection{Results}\label{sec:results}

\subsubsection{Match Outcomes}\label{sec:match-outcomes}

\Cref{tab:match-results} summarizes game outcomes across both experiments. Maia2+Psyche shows a clear monotonic score gradient across psyche zones (30.1\% stress, 44.4\% neutral, 50.8\% overconfident). Ailed-60k outcomes are near-flat across conditions (9.2\%, 9.9\%, 9.8\%), consistent with the model's overall weakness at this training scale rather than psyche-driven score variation; the signal chain's differentiating effect is concentrated in agreement and entropy metrics rather than competitive outcomes for this model.

\begin{table}[ht]
\centering
\small
\caption{Match results across experiments and controls. Score is from the psyche-modulated side's perspective against vanilla Maia2-1100. Controls use Maia2+Psyche (Experiment~B's base) with chain disabled.}
\label{tab:match-results}
\begin{tabular}{@{}llrrrrrr@{}}
\toprule
Experiment & Condition & Games & W & D & L & Score & Avg Ply \\
\midrule
\multirow{3}{*}{A: Ailed-60k} & Stress & 667 & 22 & 79 & 566 & 9.2\% & 62 \\
 & Neutral & 667 & 36 & 60 & 571 & 9.9\% & 61 \\
 & Overconfident & 667 & 33 & 65 & 569 & 9.8\% & 61 \\
\midrule
\multirow{3}{*}{B: Maia2+Psyche} & Stress & 667 & 123 & 155 & 389 & 30.1\% & 68 \\
 & Neutral & 667 & 95 & 402 & 170 & 44.4\% & 39 \\
 & Overconfident & 667 & 41 & 596 & 30 & 50.8\% & 16 \\
\midrule
\multirow{3}{*}{Control: Flat (no chain)} & Stress & 667 & 79 & 328 & 260 & 36.4\% & -- \\
 & Neutral & 667 & 93 & 327 & 247 & 38.5\% & -- \\
 & Overconfident & 667 & 88 & 317 & 262 & 37.0\% & -- \\
\midrule
\multirow{3}{*}{Control: Temp-only ($T{\approx}1.0$)} & Stress & 667 & 76 & 305 & 286 & 34.3\% & -- \\
 & Neutral & 667 & 85 & 342 & 240 & 38.4\% & -- \\
 & Overconfident & 667 & 99 & 309 & 259 & 38.0\% & -- \\
\bottomrule
\end{tabular}
\end{table}

\paragraph{Key observations.} (1)~Maia2+Psyche shows the clearest gradient: overconfident play scores 50.8\% with 596/667 draws, while stress degrades to 30.1\% with 389~losses. When applied to strong probabilities, the signal chain produces substantial behavioral variation. (2)~Ailed-60k shows a flatter gradient (9.2--9.9\%), reflecting its training data disadvantage rather than a signal chain limitation. (3)~Under overconfidence, the compressed chain essentially reproduces Maia2's top move---the personality system correctly becomes transparent at high confidence. (4)~Both controls produce flat score distributions (${\approx}$36--39\%) with no psyche-zone differentiation, confirming that the gradients are attributable to the signal chain.

\paragraph{Statistical significance and effect size.} The W/D/L distribution across psyche zones is highly significant ($\chi^2 = 628.0$, Cram\'er's $V = 0.396$, $p = 1.4 \times 10^{-134}$, Pearson chi-squared test of independence)---a medium-to-large association effect. Pairwise score comparisons confirm the gradient: stress vs.\ overconfident competitive score yields Mann-Whitney $p = 2.4 \times 10^{-43}$, $r = 0.388$ (medium effect). Confidence differences are similarly robust: all pairwise comparisons yield $p < 10^{-50}$, $r = 0.253$. The agreement spread between overconfident ($\mu = 74.7\%$, $\sigma = 17.9$\,pp, $n = 667$ games) and stress ($\mu = 42.8\%$, $\sigma = 15.7$\,pp) conditions yields Cohen's $d = (\mu_o - \mu_s)/\sigma_p = 31.9/16.84 = 1.89$, where $\sigma_p = \sqrt{(\sigma_s^2 + \sigma_o^2)/2} = 16.84$\,pp is the pooled standard deviation---a very large effect ($d > 0.8$ threshold). The flat control produces a near-uniform score of ${\approx}$37\% across all conditions, confirming that the gradient is attributable to the signal chain rather than game-state selection.

\paragraph{Behavioral signatures.} The three psyche conditions produce qualitatively distinct play profiles in Maia2+Psyche, each with recognizable human parallels:
\begin{itemize}[nosep]
  \item \textbf{Stressed} ($\psi_0 = -80$): High-variance play. The expanded distribution produces 389~losses but also 123~wins---more wins than any other condition. The wins come from unconventional move choices that occasionally catch the opponent off guard, mirroring how anxious human players sometimes stumble into surprising victories through erratic but unpredictable play.
  \item \textbf{Neutral} ($\psi_0 = 0$): Balanced baseline. The 402~draws reflect consistent, middle-of-the-road play without the extremes of either stress or overconfidence. This is the engine's natural mode: neither reckless nor rigid.
  \item \textbf{Overconfident} ($\psi_0 = +80$): Ultra-conservative. With 596/667 draws (89\%) and only 30~losses, the compressed distribution rigidly selects near-top moves. The engine rarely loses but rarely wins---it plays safe, repeating patterns rather than taking risks. This mirrors the human tendency to ``protect a lead'' by avoiding complexity, which paradoxically reduces winning chances.
\end{itemize}

\paragraph{Stochastic-opponent robustness.} A key concern is whether the draw-rate gradient is an artifact of the deterministic top-1 opponent. To test this, all six Experiment~D configurations were re-run with Maia2 sampled at temperature $T = 1.0$ (stochastic) rather than top-1 deterministic. Agreement values are unchanged (full chain: 41.9\%$\to$60.1\%$\to$76.6\%, identical to \Cref{tab:stage-ablation}), confirming that agreement depends solely on Ailed's move selection, not the opponent's policy. The draw gradient is also preserved under stochastic conditions (22.5\%$\to$62.0\%$\to$94.3\% for the full chain vs.\ 23.2\%$\to$60.3\%$\to$89.4\% under deterministic), confirming that the behavioral gradient is not an artifact of opponent determinism. The flat-preset control under stochastic conditions yields ${\approx}48\%$ draws uniformly across all zones---the psyche modulation, not game-state selection, produces the gradient.

\subsubsection{Psyche Trajectories}\label{sec:psyche-trajectories}

\Cref{fig:psyche-trajectories} shows the average psyche evolution over move number across all 667~games per condition. Several patterns emerge:

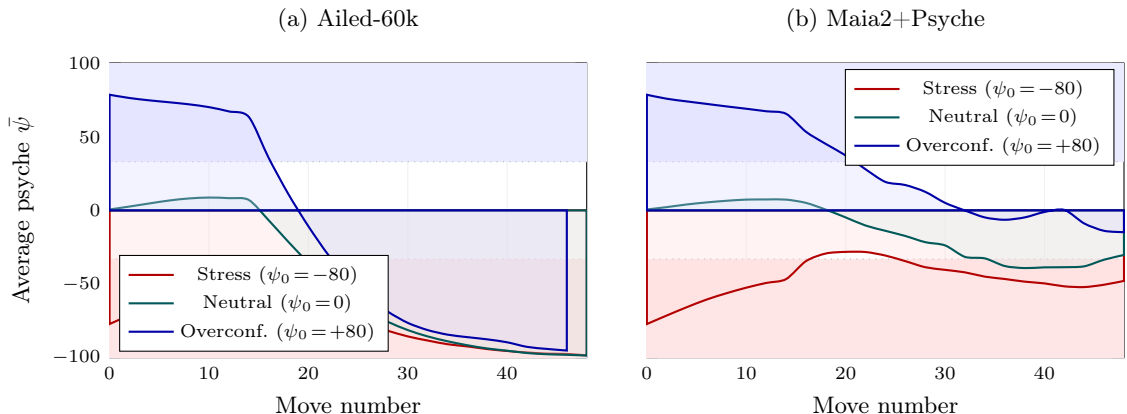
\begin{figure}[ht]
\centering
\begin{tikzpicture}
\begin{axis}[
    name=plot1,
    width=0.48\textwidth,
    height=5.5cm,
    xlabel={Move number},
    ylabel={Average psyche $\bar\psi$},
    xmin=0, xmax=48,
    ymin=-100, ymax=100,
    legend style={font=\scriptsize, at={(0.02,0.02)}, anchor=south west},
    grid=major,
    grid style={gray!12},
    tick label style={font=\scriptsize},
    label style={font=\small},
    title={\small (a) Ailed-60k},
    title style={at={(0.5,1.02)}},
]
% Psyche zone backgrounds
\fill[red!10]  (axis cs:0,-100) rectangle (axis cs:48,-33);
\fill[blue!8]  (axis cs:0, 33) rectangle (axis cs:48,100);
% Zone boundary lines
\addplot[thin, dotted, gray!60, forget plot] coordinates {(0,-33) (48,-33)};
\addplot[thin, dotted, gray!60, forget plot] coordinates {(0, 33) (48, 33)};
% Zero line
\addplot[thin, dashed, black!40, forget plot] coordinates {(0,0) (48,0)};
% Stress
\addplot[smooth, thick, red!70!black, fill=red!15, fill opacity=0.30]
    coordinates {(0,-77.2) (2,-71.2) (4,-65.3) (6,-60.1) (8,-55.8) (10,-52.7) (12,-50.7) (14,-49.7) (16,-50.1) (18,-54.0) (20,-59.4) (22,-66.0) (24,-72.2) (26,-77.8) (28,-81.8) (30,-85.6) (32,-88.4) (34,-90.9) (36,-92.4) (38,-94.2) (40,-95.5) (42,-96.5) (44,-97.3) (46,-97.5) (48,-98.5)} \closedcycle;
\addlegendentry{Stress ($\psi_0\!=\!{-80}$)}
% Neutral
\addplot[smooth, thick, teal!70!black, fill=teal!15, fill opacity=0.30]
    coordinates {(0,0.6) (2,2.6) (4,4.7) (6,6.7) (8,8.1) (10,8.7) (12,8.3) (14,7.1) (16,-6.7) (18,-21.5) (20,-35.6) (22,-48.6) (24,-59.8) (26,-69.1) (28,-75.7) (30,-81.3) (32,-85.6) (34,-88.9) (36,-91.6) (38,-93.8) (40,-95.3) (42,-96.9) (44,-97.7) (46,-98.1) (48,-98.6)} \closedcycle;
\addlegendentry{Neutral ($\psi_0\!=\!0$)}
% Overconfident
\addplot[smooth, thick, blue!65!black, fill=blue!15, fill opacity=0.30]
    coordinates {(0,78.5) (2,76.3) (4,74.7) (6,73.4) (8,72.0) (10,70.1) (12,67.2) (14,63.5) (16,35.7) (18,10.9) (20,-10.8) (22,-30.2) (24,-46.0) (26,-58.7) (28,-68.8) (30,-76.4) (32,-81.3) (34,-84.4) (36,-86.1) (38,-87.6) (40,-89.6) (42,-92.7) (44,-94.1) (46,-95.2)} \closedcycle;
\addlegendentry{Overconf.\ ($\psi_0\!=\!{+80}$)}
\end{axis}
% Coordinates from scripts/plot_psyche_trajectories.py (data-derived)
% Data source: data/paper-maia-psyche-2000/maia_psyche_metrics.json
\begin{axis}[
    at={(plot1.east)},
    anchor=west,
    xshift=0.8cm,
    width=0.48\textwidth,
    height=5.5cm,
    xlabel={Move number},
    xmin=0, xmax=48,
    ymin=-100, ymax=100,
    legend style={font=\scriptsize, at={(0.98,0.98)}, anchor=north east},
    grid=major,
    grid style={gray!12},
    tick label style={font=\scriptsize},
    label style={font=\small},
    title={\small (b) Maia2+Psyche},
    title style={at={(0.5,1.02)}},
    yticklabels={},
]
% Psyche zone backgrounds
\fill[red!10]  (axis cs:0,-100) rectangle (axis cs:48,-33);
\fill[blue!8]  (axis cs:0, 33) rectangle (axis cs:48,100);
% Zone boundary lines
\addplot[thin, dotted, gray!60, forget plot] coordinates {(0,-33) (48,-33)};
\addplot[thin, dotted, gray!60, forget plot] coordinates {(0, 33) (48, 33)};
% Zero line
\addplot[thin, dashed, black!40, forget plot] coordinates {(0,0) (48,0)};
% Stress
\addplot[smooth, thick, red!70!black, fill=red!15, fill opacity=0.30]
    coordinates {(0,-77.2) (2,-71.4) (4,-65.8) (6,-60.6) (8,-56.3) (10,-52.5) (12,-49.3) (14,-46.7) (16,-34.5) (18,-29.3) (20,-28.2) (22,-28.3) (24,-30.7) (26,-34.5) (28,-38.6) (30,-40.4) (32,-42.1) (34,-45.0) (36,-46.6) (38,-48.4) (40,-49.6) (42,-51.6) (44,-52.1) (46,-50.5) (48,-47.9)} \closedcycle;
\addlegendentry{Stress ($\psi_0\!=\!{-80}$)}
% Neutral
\addplot[smooth, thick, teal!70!black, fill=teal!15, fill opacity=0.30]
    coordinates {(0,0.6) (2,2.2) (4,4.0) (6,5.4) (8,6.5) (10,7.3) (12,7.4) (14,7.3) (16,5.1) (18,0.5) (20,-4.8) (22,-10.3) (24,-13.9) (26,-17.4) (28,-21.6) (30,-23.8) (32,-31.6) (34,-32.6) (36,-37.8) (38,-39.2) (40,-38.7) (42,-38.6) (44,-38.0) (46,-33.5) (48,-30.3)} \closedcycle;
\addlegendentry{Neutral ($\psi_0\!=\!0$)}
% Overconfident
\addplot[smooth, thick, blue!65!black, fill=blue!15, fill opacity=0.30]
    coordinates {(0,78.5) (2,75.8) (4,74.0) (6,72.3) (8,70.6) (10,69.0) (12,67.3) (14,65.6) (16,53.4) (18,45.1) (20,37.3) (22,27.9) (24,19.3) (26,17.2) (28,12.6) (30,5.1) (32,-0.1) (34,-4.8) (36,-6.4) (38,-4.8) (40,-1.1) (42,0.3) (44,-8.9) (46,-13.9) (48,-14.8)} \closedcycle;
\addlegendentry{Overconf.\ ($\psi_0\!=\!{+80}$)}
\end{axis}
\end{tikzpicture}
\caption{Average psyche trajectories across 667~games per condition. Background shading marks the three psyche zones: light red ($\psi<-33$, stress), white ($|\psi|\le 33$, neutral), light blue ($\psi>33$, overconfident); dotted lines at $\psi=\pm 33$ mark zone boundaries. Shaded curve areas show the psyche footprint relative to neutral. (a)~Ailed-60k: all conditions converge toward $\psi\approx -100$ by move~30; meaningful three-way separation is confined to the first ${\sim}$15~moves. (b)~Maia2+Psyche: the stronger model sustains separation throughout---stress stabilises around $-40$, neutral around $-30$, and overconfident decays gradually from $+80$ without collapsing. The initial forced psyche is visible at move~0 in both panels.}
\label{fig:psyche-trajectories}
\end{figure}

\paragraph{Trajectory analysis.} For Ailed-60k, the weak model's consistent losses drive all three conditions toward $\psi_{\min}$ by move~${\sim}$30, compressing the separation window to the opening phase and explaining the narrow score gradient. For Maia2+Psyche, the stronger model sustains meaningful separation throughout: stress stabilizes around $\psi \approx -40$, neutral around $-30$, and overconfident gradually decays from $+80$ toward zero. The wider separation window accounts for the dramatic score gradient (30.1--50.8\%). Both engines show the forced initialization at move~0, followed by natural evolution driven by the five positional factors.

\subsubsection{Signal Chain Auto-Tuning: A Single-Game Case Study}\label{sec:eq-autotune}

\begin{figure}[p]
\centering
\begin{tikzpicture}
% Panel (a): Psyche trajectory
\begin{axis}[
    name=panelA,
    width=0.92\textwidth,
    height=3.8cm,
    ylabel={Psyche $\psi$},
    xmin=0, xmax=54,
    ymin=-100, ymax=100,
    grid=major,
    grid style={gray!15},
    tick label style={font=\scriptsize},
    label style={font=\small},
    title={\small (a) Psyche trajectory},
    title style={at={(0.5,1.05)}},
    xticklabels={},
    extra y ticks={-80,0,80},
    extra y tick style={grid=major, grid style={dashed, gray!30}},
    extra y tick labels={},
]
\addplot[very thick, black, mark=none] coordinates {(0,-77.4) (2,-71.4) (4,-64.7) (6,-58.1) (8,-51.5) (10,-44.9) (12,-38.8) (14,-32.7) (16,2.4) (18,23.9) (20,41.2) (22,58.9) (24,67.0) (26,71.7) (28,64.0) (30,15.8) (32,-14.7) (34,-30.0) (36,-48.6) (38,-65.0) (40,-76.4) (42,-84.1) (44,-89.3) (46,-92.8) (48,-95.1) (50,-96.7) (52,-97.8) (53,-98.2)};
% Zone shading labels
\node[font=\tiny, red!70!black] at (axis cs:4,-85) {stress};
\node[font=\tiny, gray] at (axis cs:16,15) {neutral};
\node[font=\tiny, blue!70!black] at (axis cs:26,85) {overconf.};
\node[font=\tiny, red!70!black] at (axis cs:46,-85) {tilt};
% Peak annotation
\draw[-{Stealth[length=3pt]}, thin, blue!60!black] (axis cs:29,85) -- (axis cs:27,76);
\node[font=\tiny, blue!60!black] at (axis cs:33,88) {peak $\psi\!=\!73$};
\end{axis}

% Panel (b): EQ band gains — deviation from unity (gain − 1)
% Positive = boost, negative = cut; Mild is anchored at 0 throughout
\begin{axis}[
    name=panelB,
    at={(panelA.below south west)},
    anchor=north west,
    yshift=-0.3cm,
    width=0.92\textwidth,
    height=4.8cm,
    ylabel={EQ gain$\,-\,$1},
    xmin=0, xmax=54,
    ymin=-0.57, ymax=0.43,
    ytick={-0.50,-0.25,0,0.25},
    legend style={font=\tiny, at={(0.98,0.98)}, anchor=north east, legend columns=5},
    grid=major,
    grid style={gray!15},
    tick label style={font=\scriptsize},
    label style={font=\small},
    title={\small (b) Equalizer band gains --- deviation from unity (positive\,$=$\,boost; negative\,$=$\,cut)},
    title style={at={(0.5,1.05)}},
    xticklabels={},
]
% Zero baseline / Mild (anchored at 0)
\addplot[gray!55, dashed, thick, forget plot] coordinates {(0,0) (53,0)};
% Best band (green)
\addplot[smooth, thick, green!60!black, mark=none] coordinates {
    (0,-0.09)(2,-0.06)(4,-0.02)(6,0.01)(8,0.04)(10,0.08)(12,0.11)(14,0.14)
    (16,0.29)(18,0.23)(20,0.18)(22,0.12)(24,0.10)(26,0.08)(28,0.11)
    (30,0.25)(32,0.23)(34,0.15)(36,0.06)(38,-0.03)(40,-0.08)(42,-0.12)
    (44,-0.15)(46,-0.16)(48,-0.18)(50,-0.18)(52,-0.19)(53,-0.19)};
\addlegendentry{Best}
% Good band (blue) — stays positive throughout
\addplot[smooth, thick, blue, mark=none] coordinates {
    (0,0.28)(2,0.27)(4,0.26)(6,0.26)(8,0.25)(10,0.24)(12,0.24)(14,0.23)
    (16,0.20)(18,0.15)(20,0.12)(22,0.08)(24,0.07)(26,0.06)(28,0.07)
    (30,0.17)(32,0.21)(34,0.23)(36,0.25)(38,0.27)(40,0.28)(42,0.28)
    (44,0.29)(46,0.29)(48,0.30)(50,0.30)(52,0.30)(53,0.30)};
\addlegendentry{Good}
% Mild band (anchored at zero — coincides with baseline)
\addlegendimage{gray!55, dashed, thick}
\addlegendentry{Mild ($=0$)}
% Bad band (orange)
\addplot[smooth, thick, orange!80!black, mark=none] coordinates {
    (0,0.24)(2,0.20)(4,0.15)(6,0.11)(8,0.06)(10,0.01)(12,-0.03)(14,-0.07)
    (16,-0.29)(18,-0.23)(20,-0.18)(22,-0.12)(24,-0.10)(26,-0.08)(28,-0.11)
    (30,-0.25)(32,-0.20)(34,-0.09)(36,0.04)(38,0.16)(40,0.24)(42,0.29)
    (44,0.33)(46,0.35)(48,0.37)(50,0.38)(52,0.38)(53,0.39)};
\addlegendentry{Bad}
% Worst band (red)
\addplot[smooth, thick, red, mark=none] coordinates {
    (0,0.04)(2,0.00)(4,-0.05)(6,-0.09)(8,-0.14)(10,-0.19)(12,-0.23)(14,-0.27)
    (16,-0.49)(18,-0.38)(20,-0.29)(22,-0.21)(24,-0.17)(26,-0.14)(28,-0.18)
    (30,-0.42)(32,-0.40)(34,-0.29)(36,-0.16)(38,-0.04)(40,0.04)(42,0.09)
    (44,0.13)(46,0.15)(48,0.17)(50,0.18)(52,0.18)(53,0.19)};
\addlegendentry{Worst}
% Peak annotation
\draw[thin, dotted, gray] (axis cs:27,-0.57) -- (axis cs:27,0.43);
\node[font=\tiny, gray, rotate=90] at (axis cs:28.5,-0.47) {peak};
\end{axis}

% Panel (c): Signal chain parameters
\begin{axis}[
    name=panelC,
    at={(panelB.below south west)},
    anchor=north west,
    yshift=-0.3cm,
    width=0.92\textwidth,
    height=4.2cm,
    xlabel={Move number (Ailed's turns)},
    ylabel={Parameter value},
    xmin=0, xmax=54,
    ymin=0, ymax=2.1,
    legend style={font=\tiny, at={(0.02,0.98)}, anchor=north west},
    grid=major,
    grid style={gray!15},
    tick label style={font=\scriptsize},
    label style={font=\small},
    title={\small (c) Gate, dynamics, and saturation},
    title style={at={(0.5,1.05)}},
]
% Gate (scaled ×20)
\addplot[thick, blue, mark=none] coordinates {(0,0.16) (2,0.18) (4,0.20) (6,0.22) (8,0.24) (10,0.26) (12,0.28) (14,0.30) (16,0.42) (18,0.60) (20,0.72) (22,0.88) (24,0.94) (26,0.98) (28,0.92) (30,0.52) (32,0.36) (34,0.32) (36,0.26) (38,0.20) (40,0.18) (42,0.14) (44,0.14) (46,0.12) (48,0.12) (50,0.10) (52,0.10) (53,0.10)};
\addlegendentry{Gate $\tau_g \times 20$}
% Dynamics
\addplot[thick, red, mark=none] coordinates {(0,0.613) (2,0.643) (4,0.677) (6,0.709) (8,0.743) (10,0.775) (12,0.806) (14,0.836) (16,1.024) (18,1.239) (20,1.412) (22,1.589) (24,1.670) (26,1.717) (28,1.640) (30,1.158) (32,0.926) (34,0.850) (36,0.757) (38,0.675) (40,0.618) (42,0.579) (44,0.554) (46,0.536) (48,0.524) (50,0.516) (52,0.511) (53,0.509)};
\addlegendentry{Dynamics $\alpha$}
% Saturation
\addplot[thick, orange, mark=none] coordinates {(0,0.345) (2,0.357) (4,0.371) (6,0.384) (8,0.397) (10,0.410) (12,0.422) (14,0.435) (16,0.508) (18,0.584) (20,0.644) (22,0.706) (24,0.734) (26,0.751) (28,0.724) (30,0.555) (32,0.471) (34,0.440) (36,0.403) (38,0.370) (40,0.347) (42,0.332) (44,0.321) (46,0.314) (48,0.310) (50,0.307) (52,0.304) (53,0.304)};
\addlegendentry{Saturation $\sigma$}
% Unity reference
\addplot[thin, dashed, black!40] coordinates {(0,1.0) (53,1.0)};
% Peak line
\draw[thin, dotted, gray] (axis cs:27,0) -- (axis cs:27,2.1);
\end{axis}
\end{tikzpicture}
\caption{Signal chain auto-tuning during a single stress game (Ailed-60k Game~502, $\psi_0 = -80$, 54~moves, draw). \textbf{(a)}~Psyche trajectory: forced stress ($-77$) recovers through neutral into overconfidence (peak $+73$ at move~27), then collapses back to $-98$ as the position deteriorates. \textbf{(b)}~EQ band gains shown as deviation from unity (positive$=$boost, negative$=$cut). At low $\psi$ (moves 0--14), ``good'' is strongly boosted ($+0.28$) while ``worst'' is cut ($-0.27$), and ``best'' is suppressed---a diffuse selection profile. At the overconfident peak (move~27), all deviations shrink toward zero---the EQ becomes transparent. In the degradation phase (moves~30+), the pattern resembles the low-$\psi$ opening. \textbf{(c)}~Signal chain parameters track the psyche: dynamics power crosses unity at neutral (expander below, compressor above), gate opens under stress and tightens at confidence, saturation ceiling rises and falls with psyche.}
\label{fig:eq-autotune}
\end{figure}
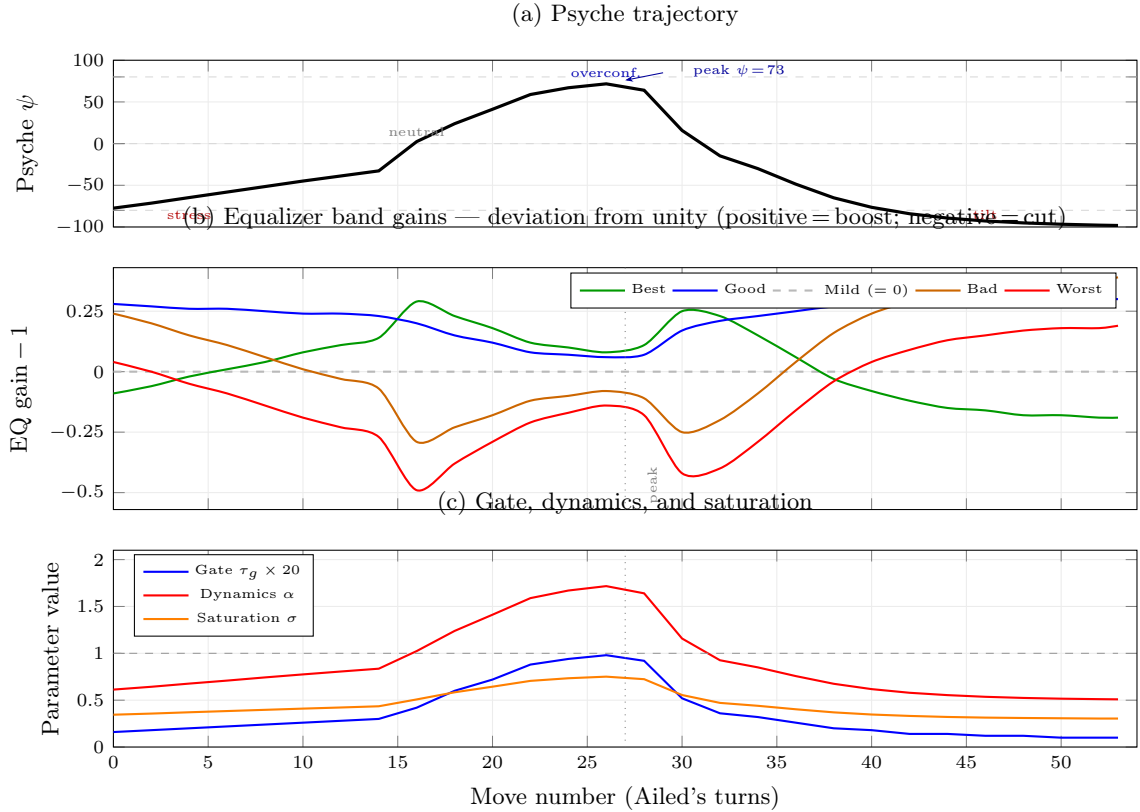

To illustrate how the signal chain adapts in real time, \Cref{fig:eq-autotune} traces a single Ailed-60k stress game (Game~502: $\psi_0 = -80$, 54~moves, draw) through its complete emotional arc. The game exhibits the classic stress-recovery-tilt pattern: starting under forced stress ($\psi = -77$), gradually recovering through neutral into overconfidence ($\psi = +73$ at move~27), then collapsing back to deep stress ($\psi = -98$) as the position deteriorates.

\paragraph{Interpretation.} The three panels show how a single scalar orchestrates the entire signal chain in real time. During the low-$\psi$ opening (moves~0--14), the open gate, expander dynamics, and low saturation ceiling collectively flatten the distribution, producing diffuse move selection. At peak overconfidence (move~27), tightened parameters sharpen it into near-deterministic play. The degradation phase (moves~30+) resembles the opening as EQ bands return to the low-$\psi$ shape, dynamics flatten, and the gate reopens. None of this requires explicit mode switching---the psyche value alone drives all transforms through the linear interpolation of \Cref{eq:lerp-psyche}.

\subsubsection{Signal Chain Behavioral Metrics}\label{sec:signal-metrics}

\Cref{tab:signal-metrics} presents the core behavioral metrics. The three key measurements are: \textbf{entropy} (Shannon entropy of the pre-chain distribution), \textbf{confidence} ($c_H = 1 - H/H_{\max}$), and \textbf{top-move agreement} (percentage of positions where the signal chain selects the engine's argmax move).

\begin{table}[ht]
\centering
\small
\caption{Signal chain behavioral metrics across conditions (per-game means; 95\% bootstrap CI for controls). Moves = total modulated-side moves; ``--'' indicates move counts not recorded for control runs.}
\label{tab:signal-metrics}
\begin{tabular}{@{}llrrrrr@{}}
\toprule
Experiment & Condition & Moves & Entropy & Confidence & Top Agree\% \\
\midrule
\multirow{3}{*}{A: Ailed-60k} & Stress & 20{,}612 & 1.610 & 0.459 & 37.1 [36.3, 37.9] \\
 & Neutral & 20{,}086 & 1.639 & 0.466 & 45.9 [45.2, 46.6] \\
 & Overconfident & 20{,}337 & 1.664 & 0.469 & 57.0 [56.4, 57.6] \\
\midrule
\multirow{3}{*}{B: Maia2+Psyche} & Stress & 22{,}606 & 1.612 & 0.497 & 41.2 [40.5, 41.9] \\
 & Neutral & 12{,}901 & 1.542 & 0.529 & 51.8 [51.0, 52.6] \\
 & Overconfident & 5{,}180 & 1.515 & 0.575 & 66.0 [65.0, 67.0] \\
\midrule
\multirow{3}{*}{Control: Flat (no chain)} & Stress & -- & 1.592 & 0.464 & 56.1 [54.7, 57.6] \\
 & Neutral & -- & 1.584 & 0.472 & 56.1 [54.7, 57.5] \\
 & Overconfident & -- & 1.595 & 0.466 & 56.4 [55.0, 57.8] \\
\midrule
\multirow{3}{*}{Control: Temp-only ($T{\approx}1.0$)} & Stress & -- & 1.570 & 0.469 & 56.5 [55.1, 57.9] \\
 & Neutral & -- & 1.565 & 0.474 & 59.0 [57.6, 60.4] \\
 & Overconfident & -- & 1.575 & 0.472 & 56.2 [54.9, 57.6] \\
\bottomrule
\end{tabular}
\end{table}

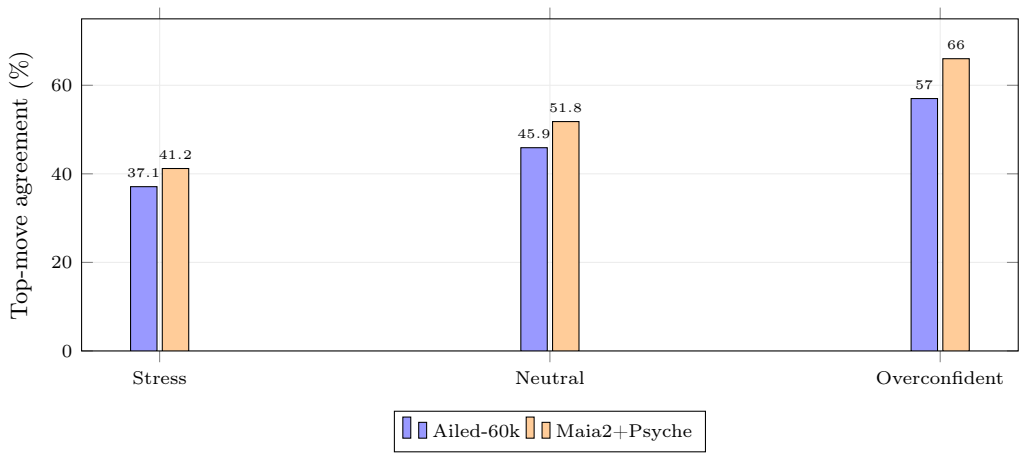
\begin{figure}[ht]
\centering
\begin{tikzpicture}
\begin{axis}[
    width=0.85\textwidth,
    height=6cm,
    ybar,
    bar width=10pt,
    ymin=0, ymax=75,
    ylabel={Top-move agreement (\%)},
    symbolic x coords={Stress,Neutral,Overconfident},
    xtick=data,
    legend style={font=\scriptsize, at={(0.5,-0.18)}, anchor=north, legend columns=2},
    grid=major,
    grid style={gray!15},
    tick label style={font=\scriptsize},
    label style={font=\small},
    nodes near coords,
    nodes near coords style={font=\tiny},
]
\addplot[fill=blue!40] coordinates {(Stress,37.1) (Neutral,45.9) (Overconfident,57.0)};
\addlegendentry{Ailed-60k}
\addplot[fill=orange!40] coordinates {(Stress,41.2) (Neutral,51.8) (Overconfident,66.0)};
\addlegendentry{Maia2+Psyche}
\end{axis}
\end{tikzpicture}
\caption{Top-move agreement by condition across both experiments (4{,}002 games). Both engines show the same monotonic gradient: stress reduces agreement (more deviation from model's top pick), overconfidence increases it (near-deterministic selection). The gradient is steeper for Maia2+Psyche because its stronger model probabilities give the signal chain more concentrated input to work with.}
\label{fig:agreement-both}
\end{figure}

\paragraph{Key findings.} Both engines exhibit a clear monotonic gradient: stress $<$ neutral $<$ overconfident. For Ailed-60k the spread is 19.9~pp ($37.1\% \to 57.0\%$); for Maia2+Psyche it is 24.8~pp ($41.2\% \to 66.0\%$). Confidence increases monotonically across conditions in both experiments. Maia2+Psyche under overconfidence produces 66\% agreement with vanilla Maia2's top move, confirming the chain becomes nearly transparent at high psyche. Move counts differ sharply for Maia2+Psyche (22{,}606 stress vs.\ 5{,}180 overconfident) because overconfident games are short---mostly threefold repetition draws within $\sim$16 plies. Both controls show near-zero agreement spread (${\leq}$2.8\,pp), providing a clean null baseline.

\subsubsection{Cross-Engine Comparison}\label{sec:cross-engine}

The parallel experimental design reveals that the signal chain's behavioral gradient is \emph{architecture-agnostic}. \Cref{tab:gradient-comparison} compares the stress-to-overconfident spread for each metric.

\begin{table}[ht]
\centering
\small
\caption{Signal chain gradient comparison: stress $\to$ overconfident spread across engines.}
\label{tab:gradient-comparison}
\begin{tabular}{@{}lrr@{}}
\toprule
Metric & Ailed-60k spread & Maia2+Psyche spread \\
\midrule
Score & $+0.6$~pp (9.2 $\to$ 9.8) & $+20.7$~pp (30.1 $\to$ 50.8) \\
Top agree\% & $+19.9$~pp (37.1 $\to$ 57.0) & $+24.8$~pp (41.2 $\to$ 66.0) \\
Confidence & $+0.010$ (0.459 $\to$ 0.469) & $+0.078$ (0.497 $\to$ 0.575) \\
Entropy & $+0.054$ (1.610 $\to$ 1.664) & $-0.097$ (1.612 $\to$ 1.515) \\
\bottomrule
\end{tabular}
\end{table}

\paragraph{Interpretation.} Top-move agreement is the cleanest metric for the signal chain: it measures how much the chain deviates from the engine's preferred move under different psyche states. Both engines show ${\sim}$20--25~pp spreads, confirming consistent operation regardless of model strength. The score gradient is compressed for Ailed-60k because its weak model loses most games regardless of psyche; for Maia2+Psyche, the score gradient materializes fully because the model is competitive with the opponent.

% ----------------------------------------------------------------------------
% EXPERIMENT C: OUTCOME PREDICTOR ABLATION
% ----------------------------------------------------------------------------
\subsection{Experiment~C: Outcome Predictor Ablation}\label{sec:ablation}

The initial system combined entropy confidence and outcome confidence from a dedicated 10.7M-parameter predictor via geometric mean $c = \sqrt{c_\text{entropy} \cdot c_\text{outcome}}$. Since entropy already captures decision certainty from the move distribution, the question is whether the outcome predictor meaningfully contributes to the behavioral gradient.

\paragraph{Design.} I replay both engine configurations (Ailed-60k, Maia2+Psyche) under three confidence modes:
\begin{itemize}[nosep]
  \item \textbf{Full} (baseline): $c = \sqrt{c_\text{entropy} \cdot c_\text{outcome}}$ --- geometric mean of both signals.
  \item \textbf{Entropy-only}: $c = c_\text{entropy}$ --- outcome predictor loaded but excluded from confidence.
  \item \textbf{Fixed}: $c = 1.0$ --- EQ always fully engaged, no confidence gating.
\end{itemize}
Each ablation mode plays 600~games (200 per psyche condition, balanced colors) against vanilla Maia2-1100, totaling 2{,}400 new games. The \texttt{full} baselines are Experiments~A and~B (2{,}001 games each).

\paragraph{Results.} \Cref{tab:ablation} summarizes the ablation across both engines. The key metric is the top-move agreement spread (overconfident minus stress), which measures the signal chain's behavioral gradient.

\begin{table}[ht]
\centering
\small
\caption{Outcome predictor ablation: top-move agreement (\%) and spread across confidence modes. Spread = overconfident $-$ stress.}
\label{tab:ablation}
\begin{tabular}{@{}llrrrrr@{}}
\toprule
Engine & Mode & Stress & Neutral & Overconf. & Spread & Score\textsubscript{s$\to$o} \\
\midrule
Ailed-60k & Full & 37.1 & 45.9 & 57.0 & 19.9\,pp & 9.2--9.8\% \\
Ailed-60k & Entropy-only & 37.4 & 45.8 & 56.3 & 18.9\,pp & 10.0--5.5\% \\
Ailed-60k & Fixed ($c{=}1$) & 36.5 & 44.2 & 55.4 & 18.9\,pp & 10.2--9.0\% \\
\midrule
Maia2+Psyche & Full & 41.2 & 51.8 & 66.0 & 24.8\,pp & 30.1--50.8\% \\
Maia2+Psyche & Entropy-only & 43.2 & 51.5 & 71.3 & 28.1\,pp & 25.8--52.0\% \\
Maia2+Psyche & Fixed ($c{=}1$) & 41.1 & 50.1 & 71.6 & 30.5\,pp & 27.8--50.5\% \\
\bottomrule
\end{tabular}
\end{table}

\paragraph{Findings.}
\begin{enumerate}[nosep]
\item \textbf{The outcome predictor is not essential for the behavioral gradient.} For Ailed-60k, the agreement spread is virtually identical across all three modes (19.9 vs.\ 18.9 vs.\ 18.9\,pp). Removing the outcome predictor changes the gradient by less than 1\,pp.

\item \textbf{For strong models, removing the outcome predictor \emph{widens} the gradient.} Maia2+Psyche shows a spread increase from 24.8\,pp (full) to 28.1\,pp (entropy-only) to 30.5\,pp (fixed). The outcome predictor was partially dampening the signal chain's effect by moderating the confidence signal.

\item \textbf{Confidence values differ but behavioral effects converge.} Mean confidence varies across modes (trivially 1.0 for fixed, 0.42--0.52 for entropy-only, 0.46--0.58 for the full geometric mean), yet the agreement gradient remains similar---suggesting that gate, dynamics, and saturation carry most of the behavioral weight independently of the confidence signal.

\item \textbf{Competitive outcomes are robust to confidence mode.} Score gradients remain comparable across modes: Maia2+Psyche shows ${\sim}$20--26\,pp score spreads in all configurations, confirming that psyche-driven performance degradation is a property of the signal chain stages rather than the confidence gating.
\end{enumerate}

\paragraph{Implication.} Entropy confidence alone is sufficient for the EQ wet/dry mix. The outcome predictor's 10.7M parameters (31\% of total model size) are unnecessary for the core personality system and have been removed from the production system. The ablation results (Table~\ref{tab:ablation}) are retained as evidence that this removal does not degrade---and for strong models actually widens---the behavioral gradient.

% ----------------------------------------------------------------------------
% EXPERIMENT D: STAGE-WISE SIGNAL CHAIN ABLATION
% ----------------------------------------------------------------------------
\subsection{Experiment~D: Stage-wise Signal Chain Ablation}\label{sec:stage-ablation}

Experiment~C showed that confidence gating (outcome predictor) is not essential for the behavioral gradient. This experiment asks: \emph{which stages within the core signal chain carry the majority of the behavioral effect?} The signal chain comprises four sequential stages---gate, dynamics, EQ, and saturation---and Experiment~D isolates each stage's contribution by ablating one stage at a time from the Maia2+Psyche configuration.

\paragraph{Design.} Six configurations are evaluated, each running 334~games per psyche zone (1{,}002~games per configuration, 6{,}012~games across all configurations):
\begin{itemize}[nosep]
  \item \textbf{Full chain}: all stages active (baseline).
  \item \textbf{No gate}: gate stage bypassed; dynamics/EQ/saturation always active.
  \item \textbf{No dynamics}: dynamics stage bypassed; gate/EQ/saturation active.
  \item \textbf{No EQ}: EQ stage bypassed; gate/dynamics/saturation active.
  \item \textbf{No saturation}: saturation stage bypassed; gate/dynamics/EQ active.
  \item \textbf{No gate~+~no dynamics}: both stages removed; only EQ and saturation active.
\end{itemize}

\paragraph{Results.} \Cref{tab:stage-ablation} shows top-move agreement across all six configurations, with the flat-preset control (all gains~=~1.0, dynamics exponent~=~1.0) as the no-chain reference.

\begin{table}[ht]
\centering
\small
\caption{Stage-wise signal chain ablation: top-move agreement (\%) by psyche zone (Maia2+Psyche, 334 games per zone per config). Flat control (667 games/zone) shown for reference. Spread~=~overconfident~$-$~stress.}
\label{tab:stage-ablation}
\begin{tabular}{@{}lrrrr@{}}
\toprule
Configuration & Stress & Neutral & Overconf. & Spread \\
\midrule
Full chain      & 41.9 & 60.1 & 76.6 & 34.7\,pp \\
No gate         & 35.6 & 53.5 & 72.0 & 36.4\,pp \\
No dynamics     & 54.1 & 61.7 & 68.4 & 14.3\,pp \\
No EQ           & 44.3 & 62.5 & 74.9 & 30.6\,pp \\
No saturation   & 48.3 & 64.7 & 78.2 & 29.9\,pp \\
No gate~+~no dynamics & 51.7 & 52.9 & 54.1 & 2.4\,pp \\
\midrule
Flat control (no chain) & 56.1 & 56.1 & 56.4 & 0.3\,pp \\
\bottomrule
\end{tabular}
\end{table}

\paragraph{Findings.}
\begin{enumerate}[nosep]
\item \textbf{Dynamics is the dominant stage.} Removing dynamics collapses the stress-to-overconfident spread from 34.7\,pp to 14.3\,pp---a 59\% reduction. The dynamics stage (power-law compression of the move probability distribution with psyche-dependent exponent) directly controls how ``peaked'' or ``flat'' the output distribution is under each psyche state.

\item \textbf{Gate and dynamics together are necessary for the gradient.} The ``no gate~+~no dynamics'' configuration produces only 2.4\,pp spread---comparable to the flat-preset baseline (0.3\,pp, where the signal chain does not modulate the distribution). Removing both stages eliminates the behavioral gradient, confirming that EQ and saturation alone cannot produce meaningful zone differentiation.

\item \textbf{EQ and saturation contribute secondary amplification.} Removing either stage reduces the spread by ${\sim}$4--5\,pp (no EQ: 34.7~$\to$~30.6\,pp; no saturation: 34.7~$\to$~29.9\,pp). Both stages amplify the core dynamics effect but cannot independently produce it.

\item \textbf{Gate removal shifts the gradient's locus.} Without the gate, the stressed zone drops further (35.6\% vs.\ 41.9\% agreement) while the overconfident zone decreases slightly (72.0\% vs.\ 76.6\%), widening the spread to 36.4\,pp. The gate moderates the most extreme stress behavior while slightly boosting overconfident determinism.
\end{enumerate}

\paragraph{Implication.} The dynamics stage is the engine of psyche-driven behavioral variation. EQ, saturation, and gate amplify and refine it, but cannot independently produce a meaningful gradient. In principle, a minimal personality module could retain the dynamics stage alone---though the full chain offers richer control.

% ----------------------------------------------------------------------------
% SUMMARY
% ----------------------------------------------------------------------------
\subsection{Summary of Findings}\label{sec:exp-summary}

\begin{table}[ht]
\centering
\small
\caption{Summary of key experimental findings from 12{,}414 games across four experiments.}
\label{tab:summary}
\begin{tabular}{@{}llr@{}}
\toprule
Finding & Evidence & Ref. \\
\midrule
Signal chain produces monotonic gradient & Top agree: 37--57\% (A), 41--66\% (B) & \Cref{sec:signal-metrics} \\
Gradient is architecture-agnostic & ${\sim}$20--25~pp spread for both engines & \Cref{sec:cross-engine} \\
Overconfident = near-transparent & 66\% Maia2 agree, 596/667 draws (B) & \Cref{sec:match-outcomes} \\
Stress degrades competitive score & 30.1\% stress vs.\ 50.8\% overconfident (B) & \Cref{sec:match-outcomes} \\
Confidence tracks psyche state & Monotonic increase: 0.459 $\to$ 0.575 & \Cref{sec:signal-metrics} \\
Training data gap limits Ailed & 9.2--9.9\% score vs.\ 30--51\% for Maia2+Psyche & \Cref{sec:match-outcomes} \\
Outcome predictor not essential for gradient & $<$1\,pp spread change when removed (A) & \Cref{sec:ablation} \\
Entropy confidence sufficient for EQ mix & 18.9--30.5\,pp spreads without outcome & \Cref{sec:ablation} \\
Dynamics is the dominant signal chain stage & Removal collapses spread: 34.7~$\to$~14.3\,pp & \Cref{sec:stage-ablation} \\
Gate~+~dynamics necessary for gradient & Without both: 2.4\,pp spread (near-flat) & \Cref{sec:stage-ablation} \\
Gradient robust to opponent stochasticity & Agreement unchanged under $T{=}1.0$ opponent; $d{=}1.89$ & \Cref{sec:match-outcomes} \\
\bottomrule
\end{tabular}
\end{table}

\subsection{Reproducibility and Availability}\label{sec:reproducibility}

\textbf{Code.} Source code is available at \url{https://github.com/chrnx-dev/ailed-chess}
under the Apache-2.0 license. This paper is licensed under Creative
Commons Attribution 4.0 International (CC~BY~4.0).

\textbf{Data.} Experiments use standard-rated games from the Lichess open database
(CC0)~\citep{lichess_database}, months 2016-01 through 2016-04 and 2016-07
(five monthly archives). Raw PGN data is not redistributed; download scripts are
provided in the repository.

\textbf{Opponent.} Maia2~\citep{tang2024maia2} at ELO~1100, using top-probability
(deterministic argmax) move selection. Determinism is intentional to enable
controlled within-position comparisons; it may increase draw rates via repetition.
Maia2 is an optional dependency for evaluation only; the AILED core engine has no
dependency on it.

\textbf{Experimental setup.} Random seed: 42. Training hardware: NVIDIA RTX 3060;
evaluation hardware: Apple Silicon (arm64). Python~3.12.8, PyTorch~2.1.0,
python-chess (see \texttt{pyproject.toml}). Full version metadata is written to
\texttt{data/paper/reproducibility.json} alongside each experiment's results.

% ============================================================================
% ============================================================================
% 8. DISCUSSION
% ============================================================================
\section{Discussion}\label{sec:discussion}

\subsection{Observed Error Patterns}

The psyche model produces patterns that look a lot like what players and commentators describe as tilt and overconfidence. I want to be clear: I'm not claiming these patterns are validated against human behavior---that would take controlled studies with actual players. But the qualitative parallels are interesting enough to walk through.

\paragraph{Tilt.} When the engine drops material, the psyche goes negative. That opens the noise gate, flattens the distribution through the dynamics expander, and boosts the mid-rank EQ bands. More errors follow, which cause more losses, which push the psyche further down---a feedback loop that is structurally similar to what competitive gamers call ``tilt'' \citep{palomaki2013dont}. The clamp at $\psi_{\min} = -100$ keeps it from running off to negative infinity, and overnight decay eventually pulls it back. In the experiments, forced stress gives both engines their lowest top-move agreement: 37.1\% for Ailed-60k and 41.2\% for Maia2+Psyche.

\paragraph{Overconfidence traps.} A dominant board elevates $\psi$. The gate tightens, dynamics compress, EQ gets bypassed---and the engine's play becomes nearly deterministic. It's reminiscent of the old observation that strong players blunder most in positions they think they're winning \citep{degroot1965thought}, though I should note the mechanism here is purely computational (narrowed probability mass), not a validated model of what happens in someone's head. In Experiment~B, the extreme overconfident condition produces 596 draws out of 667 games, most of them threefold repetitions within about 8~plies. Two near-deterministic agents just cycle. That's deliberate---I pushed the parametrization to an extreme to see where the chain's boundaries are. The draw rate is the point; it shows the chain going transparent.

\paragraph{Recovery cycles.} Nothing is permanent. The daily decay (\Cref{def:decay}), with $\lambda = 0.20$ and a half-life of about 4~days, pulls even the worst tilt back to $|\psi| < 5$ within two weeks of not playing.

\subsection{Architecture-Agnostic Personality}

The dual-engine experiment tells a clear story: the signal chain produces consistent behavioral gradients (${\sim}$20--25~pp spread) even when the probability sources underneath differ by ${\sim}$2{,}800$\times$ in training data. The exact magnitude isn't the same---Ailed-60k gives 19.9~pp, Maia2+Psyche gives 24.8~pp (\Cref{tab:gradient-comparison})---and that makes sense, because a better-calibrated distribution leaves more room for psyche-driven differentiation. But the qualitative ordering (stress $<$ neutral $<$ overconfident) holds either way. The variation is coming from the chain, not from the model.

\paragraph{Outcome predictor contribution.} Experiment~C makes a practical case for simplicity. Entropy confidence alone gives nearly the same agreement spreads ($<$1\,pp difference for Ailed-60k), and for Maia2+Psyche, taking the outcome predictor \emph{out} actually \emph{widens} the gradient (24.8 $\to$ 28.1\,pp)---it was dampening the chain. So I removed it from the production system.

\subsection{Why Minimal Training Data Works}\label{sec:minimal-data}

Why does a model trained on just 60K games work at all? Because at 1100~ELO, players are surprisingly consistent. They see the same obvious threats, consider the same candidate moves, and make the same kinds of mistakes. You don't need millions of games to learn those distributional patterns---you need enough to get the shape of the distribution right. The psyche layer then leans on that distribution, deciding how confidently and how narrowly the engine acts within it. The interesting behavior comes from that interaction, not from raw predictive accuracy.

Practically, this means you can spin up the system at any ELO: train a small model on games from that band, plug in the signal chain, and you're done. The psyche and personality parameters don't need retraining---only the probability source changes. That makes the whole thing accessible without serious compute, and it opens up the idea of personalized opponents trained on a single player's own game history.

\subsection{Design Choices and Trade-offs}

\paragraph{Audio signal chain metaphor.} Why audio? Because the stages map onto decision-making under pressure in a way that felt right: gate controls how wide the attention is, dynamics controls how peaked the choice becomes, the EQ weights different quality tiers, and saturation prevents any one move from hogging all the probability. Whether the analogy is deep or just useful is an open question---but Experiment~D (\Cref{sec:stage-ablation}) at least grounds it empirically. The dynamics stage does most of the heavy lifting: removing it alone collapses the spread from 34.7\,pp to 14.3\,pp. EQ and saturation each add ${\sim}$4--5\,pp on top. And since every stage is differentiable, there's a path to learning personality parameters end-to-end down the road.

\paragraph{Bounded cognition as a design principle.} Ailed doesn't do deep search, and that's on purpose. Over the board, human players work with short tactical horizons and fragile plans. Ailed's forward-model prediction plus its thinking module reflect that. I'm not trying to replicate how engines compute optimal moves---I'm trying to produce play that \emph{feels} like someone choosing under pressure.

\paragraph{Scalar psyche.} One number for the whole emotional picture is obviously a simplification. Human affect is multidimensional \citep{russell1980circumplex}, and you could imagine a vector $\boldsymbol{\psi} \in \mathbb{R}^k$ with separate axes for stress, confidence, frustration, focus. I went with a scalar for three practical reasons: it's enough to parameterize the full signal chain via linear interpolation, the weight vector~$\mathbf{w}$ still gives you configurability, and---perhaps most importantly---a single number is easy to visualize, interpret, and debug.

\paragraph{Asymmetric disruption rates.} The choice $r^{-} = 0.80 > r^{+} = 0.60$ (\Cref{def:disruption}) is grounded in attentional control theory \citep{eysenck2007anxiety}, which holds that anxiety impairs the goal-directed attentional system while leaving the stimulus-driven system intact.

\subsection{Why Centipawn Loss Does Not Capture Psyche Effects}\label{sec:cpl-layer}

It is natural to ask whether centipawn loss (CPL)---the standard per-move quality metric in chess engine evaluation---shows the same psyche-driven gradient as the behavioral metrics reported in \Cref{sec:results}. The short answer is that it does not, and this is expected given the system's architecture.

\paragraph{Selection layer vs.\ position distribution.} The psyche system operates at the \emph{selection} layer: the MovePredictor outputs identical logits for identical board positions regardless of $\psi$---it does not enter the forward pass. What $\psi$ modulates is which move gets \emph{selected} from the resulting distribution, via the signal chain's gate, dynamics, EQ, and saturation stages. CPL measures the quality of each move played relative to the Stockfish-optimal continuation, so in principle it \emph{could} reflect selection-layer effects (choosing worse moves means higher CPL). In practice, however, psyche zones correlate with position difficulty by construction---losing material pushes $\psi$ negative, and those same positions are objectively harder to play. The CPL signal is therefore dominated by the position distribution encountered in each zone, not by the selection-layer modulation. A stressed engine and an overconfident engine examining the \emph{same} position produce the same logits, the same top move, and the same CPL for that top move---they differ only in how often they \emph{deviate} from it. But across a game, the stressed engine visits harder positions, confounding any selection-layer CPL signal.

\paragraph{The floor effect.} Preliminary CPL analysis revealed an apparent paradox: in panic zones ($\psi \leq -60$), mean CPL appeared \emph{lower} (48.4~cp) than in stressed zones (66.8~cp). This does not reflect higher-quality play under panic. Rather, it is a floor effect: positions that push the psyche into panic are already severely negative (often $-400$~cp or worse). There are simply fewer centipawns remaining to lose on a single move without triggering mate-transition filters. The CPL distribution compresses toward zero as the evaluation floor approaches, creating the illusion of improved play quality in the worst positions.

\paragraph{Human baseline: CPL correlates with position difficulty, not psyche.}
To confirm that the CPL pattern is a property of the \emph{positions} associated with each psyche zone rather than an artifact of Ailed's architecture, I computed CPL for human players in the same psyche zones. The baseline uses 7,748 Lichess standard-rated games from 2016 in the 950--1250~ELO band, evaluated at Stockfish depth~8 with 25,000~moves sampled per zone. Because the psyche metric is computed from board features visible to both sides, the same psyche zone boundaries apply: stressed positions are objectively worse, overconfident positions are objectively better.

\Cref{tab:human-ailed-cpl} shows the result. Both curves share the same shape: stressed positions carry the highest CPL (humans 96.8~cp, Ailed 66.8~cp), overconfident positions the lowest for humans (54.5~cp), and neutral positions fall between. The one reversal is in the overconfident zone, where Ailed's CPL (58.2~cp) slightly exceeds the human baseline (54.5~cp)---the compressed signal chain occasionally fixates on a near-top move that happens to be suboptimal, while human players with broader consideration avoid that particular trap. This pattern is \emph{not} caused by the psyche system---it reflects position difficulty. Positions that leave the side-to-move trailing on material, king safety, and mobility (the stressed zone) are objectively harder to play correctly, and both human players and Ailed make more errors there. Positions where the side-to-move holds a large advantage (overconfident zone) admit fewer costly mistakes simply because the winning side has more options of similar quality.

\begin{table}[ht]
\centering
\small
\caption{Mean centipawn loss per psyche zone: human 1100-ELO players (Lichess 2016, $n=7{,}748$ games, 25,000 moves/zone, Stockfish depth~8) vs.\ Ailed (2,001 experiment games, 300 moves/zone). Difference = Human $-$ Ailed.}
\label{tab:human-ailed-cpl}
\begin{tabular}{@{}lllr@{}}
\toprule
Zone & Human CPL [95\% CI] & Ailed CPL [95\% CI] & Difference \\
\midrule
Panic      & 64.9 [63.3, 66.5] & 48.4 [39.6, 58.5] & $+16.5$ \\
Stressed   & 96.8 [94.8, 98.9] & 66.8 [55.7, 78.8] & $+30.0$ \\
Neutral    & 76.8 [75.2, 78.4] & 61.3 [53.2, 70.1] & $+15.5$ \\
Confident  & 79.4 [77.9, 81.0] & 65.0 [54.1, 77.5] & $+14.4$ \\
Overconf   & 54.5 [53.4, 55.7] & 58.2 [50.1, 67.6] & $-3.7$  \\
\bottomrule
\end{tabular}
\end{table}

\paragraph{Two-level interpretation.}
The data support a two-level reading:

\textbf{Level~1 (position difficulty, shared by both).} Both human players and Ailed show CPL that tracks the inherent difficulty of the positions associated with each psyche zone. Stressed positions---where the side-to-move is behind---are harder to navigate, and both agents make more centipawn errors there. Overconfident positions---where the side-to-move leads---offer more margin for error. This Level~1 pattern would appear even in a flat engine with no psyche modulation; it is a property of the position distribution, not of the personality system.

\textbf{Level~2 (behavioral differentiation, from the EQ chain).} Despite facing the same Level~1 position difficulty as a flat engine would, Ailed with psyche modulation shows a 20-point gap in competitive score between stressed (30.1\%) and overconfident (50.8\%) conditions. A flat-temperature control (\Cref{sec:match-outcomes}, \Cref{tab:match-results}) produces ${\sim}37\%$ uniformly. This behavioral differentiation---present in win rates and agreement patterns but absent from per-move CPL---is the measurable contribution of the psyche EQ signal chain. CPL measures at the model layer, below where psyche acts; behavioral metrics measure at the selection layer, where it does.

\paragraph{The right metrics for a selection-layer intervention.} The appropriate evidence for psyche effects is behavioral: how does the \emph{pattern of selections} change across conditions? The W/D/L distribution ($\chi^2 = 628.0$, Cram\'er's $V = 0.396$, $p = 1.4 \times 10^{-134}$), competitive score gradient (30.1\% to 50.8\%), top-move agreement spread (${\sim}$20--25\,pp), and confidence gradient all measure at the selection layer where psyche operates. An analogy may help: a stressed human driver does not forget traffic rules; they change which decisions they make---accelerating when they should brake, taking risks they would normally avoid. Centipawn loss measures rule compliance on a single turn; psyche measures the pattern of decisions across a game.

\subsection{Limitations}

\begin{enumerate}
    \item \textbf{Training data asymmetry.} Ailed trains on ${\sim}$60K games versus Maia2's 169M, making overall competitive performance non-comparable. The Maia2+Psyche experiment isolates signal chain effects from model weakness; scaling Ailed's training data is the clearest path to stronger native play.
    \item \textbf{Psyche convergence under weak models.} For Ailed-60k, consistent material loss drives all three initial conditions toward $\psi_{\min}$ by move~${\sim}$30, compressing the differentiating window to the first ${\sim}$15~moves. Maia2+Psyche maintains meaningful separation throughout, suggesting that convergence rate depends on the underlying model's competitive strength. Slower factor-driven decay or position-independent psyche persistence could extend the window for weaker models.
    \item \textbf{Hand-tuned personality presets.} The six presets were designed by analogy rather than learned from data. A natural extension is training a small network to predict signal chain parameters from game histories, enabling data-driven personality discovery.
    \item \textbf{Deliberate bounded search.} Ailed does not use deep tree search by design. This aligns with its behavioral realism objective, but limits tactical depth.
    \item \textbf{Opponent diversity.} All games are against a single opponent (Maia2-1100). Stochastic-temperature replications (\Cref{sec:match-outcomes}) confirm that gradients persist independent of opponent policy, but testing against multiple ELO levels and engine types remains future work.
    \item \textbf{Move quality metrics.} The experiments use competitive outcomes, agreement, and entropy as primary evidence. Centipawn loss (CPL) was investigated but does not capture psyche effects, for reasons discussed in \Cref{sec:cpl-layer}: the psyche operates at the selection layer, not the model layer, and CPL measures positional quality below the level where psyche acts. The behavioral metrics (W/D/L distributions, agreement spread, confidence gradient) are the appropriate evidence for a selection-layer intervention.
\end{enumerate}

\subsection{Future Work}\label{sec:future-work}

The experiments in this paper validate the signal chain as a mechanism; the next phase is about validating the \emph{behavior} it produces. Several concrete directions are underway or planned, roughly in order of how soon I expect results.

\paragraph{Live play and behavioral validation.}
Ailed is deployed as \texttt{ailedbot} on Lichess, accepting challengers across all ELO ranges. The bot logs full PGNs with psyche trajectories, per-move Stockfish evaluations, and thinking mode activation traces. As discussed in \Cref{sec:cpl-layer}, CPL does not capture psyche effects because the psyche operates at the selection layer rather than the model layer. The live play data will therefore focus on behavioral metrics: W/D/L distributions, agreement patterns, and entropy profiles across psyche conditions with real opponents rather than the controlled Maia2-1100 setup. Early data collection is underway (under 100~games at the time of writing).

A secondary question is whether Ailed's behavioral signatures under stress---high-variance play, increased deviation from model top-move, occasional surprising wins---resemble the behavioral profiles of human players under pressure. This would require matching Ailed's game-level statistics against human games filtered by similar positional stress indicators, a comparison that the Lichess open database makes feasible.

\paragraph{Per-personality evaluation.}
The six personality presets (flat, classical, rock, jazz, metal, human) are defined in \Cref{tab:presets} but only ``human'' has been experimentally evaluated. A natural follow-up is to rotate through all six presets on the Lichess bot, collecting enough games per preset to characterize their behavioral signatures: does ``jazz'' really avoid the obvious move more often? Does ``metal'' produce more chaotic play? Does ``classical'' hold up better under stress than ``rock''? This rotation isn't deployed yet, but the infrastructure is in place. Even a modest dataset (200--500~games per preset) would answer whether the presets produce qualitatively distinct play or collapse into similar behavior once the psyche takes over.

\paragraph{Thinking mode and study mode validation.}
The bot already logs thinking mode activations---when plans are generated, whether the opponent's move matched the prediction, and whether the psyche disruption check fired. The study mode runs offline between sessions. Neither module has been evaluated beyond unit tests. The Lichess data will provide the first empirical look at how often plans succeed, how psyche extremes affect plan survival rates, and whether study sessions measurably improve performance in subsequent games. I expect the thinking mode data to accumulate faster than the study mode data, since the latter requires enough games to trigger the weekly fine-tuning loop.

\paragraph{Learned personality presets.}
The current presets are hand-tuned by analogy to music genres. A more principled approach would learn preset parameters from data: train a small network that takes a player's game history and outputs signal chain parameters (EQ gains, dynamics anchors, gate thresholds, saturation ceilings) that reproduce that player's behavioral patterns. I'm prototyping this, but it depends on having enough human game data with varied behavioral profiles---which circles back to the Lichess bot accumulating games across different opponents and conditions. The goal would be to show that distinct ``personalities'' emerge from data rather than from hand-tuning, and that the signal chain's parameter space is expressive enough to capture real stylistic variation.

\paragraph{Human-subject validation.}
The strongest test of the psyche model would be a controlled study where human players face Ailed under different psyche conditions (or personality presets) without knowing which condition they're in, then rate the opponent on perceived realism, engagement, and difficulty. I haven't designed this study yet, but it's on the roadmap. The practical challenge is recruiting enough players and controlling for confounds (player skill, game length, opening choice). A simpler first step might be a post-game survey on Lichess: after playing \texttt{ailedbot}, ask the opponent whether the bot felt ``human-like,'' ``frustrating,'' ``predictable,'' or ``engaging.'' This wouldn't be a rigorous study, but it would provide directional signal at low cost.

\paragraph{Beyond chess.}
The signal chain is game-agnostic by construction---it operates on probability distributions over discrete actions, with no chess-specific assumptions. In principle, any turn-based game with a move predictor could be wrapped in the same framework: Go, poker, card games, tactical RPGs. I don't have concrete plans for other games yet; the priority is getting the chess evaluation right first. But once the framework is validated with human data in chess, exploring whether the same psyche dynamics produce plausible behavior in structurally different games would be a natural next step.

\subsection{Ethical Considerations}

A chess engine that deliberately plays imperfectly raises fair-play concerns. Ailed is designed as a training partner and research platform, not for competitive deception. Its psyche system is transparent and configurable; the engine self-identifies via UCI commands and emits \texttt{info string planned} during thinking mode. A public dashboard exposes psyche state and game history. Ailed is registered as the bot account \texttt{ailedbot} on Lichess (\url{https://lichess.org/@/ailedbot}) under standard platform terms.

% ============================================================================
% 9. CONCLUSION
% ============================================================================
\section{Conclusion}\label{sec:conclusion}

This paper introduced the psyche model---a framework built around a simple idea: the variability that emotional dynamics introduce into decision-making isn't noise to be eliminated, it's structure worth formalizing. One bounded scalar, updated from positional features after each move, parameterizes a multi-stage signal chain that reshapes any engine's move distribution in real time. The main empirical takeaway is that this chain produces consistent behavioral gradients whether the probability source underneath was trained on 60K games or 169M. The variation comes from the chain, not from the model.

Experiment~D turned up something I hadn't expected going in: the dynamics stage alone carries most of the behavioral effect. EQ and saturation help, but you could strip them away and still get a meaningful gradient. A minimal personality module could, in principle, be just the dynamics stage. That said, the full chain gives you much finer-grained control, and the EQ in particular makes the system more interpretable.

Human baseline data from 7,748 Lichess 1100-ELO games confirms that CPL correlates with position difficulty uniformly across psyche zones, validating that behavioral metrics---not per-move quality---are the appropriate measure of psyche's effect.

The next phase of this work (\Cref{sec:future-work}) centers on validating the \emph{behavior} the chain produces, not just the mechanism. Ailed is already live on Lichess as \texttt{ailedbot}, collecting game data with psyche logs and per-move metrics. The immediate priorities are behavioral validation across psyche conditions using real opponents, per-personality evaluation across all six presets, and thinking mode validation from live play logs. Further out, I'm prototyping learned personality presets and considering a human-subject study to test perceived realism.

None of this is chess-specific, really. Any game AI that outputs move probabilities can be wrapped in the signal chain. The broader bet is that formalizing how an internal state variable---one informed by affect research---can shape decisions is a path toward AI opponents that feel less like calculators and more like someone across the board.

% ============================================================================
% AVAILABILITY
% ============================================================================
\section{Software Availability}

The AILED psyche and personality (EQ) system is released as open-source software on GitHub under the Apache 2.0 License:

\textbf{GitHub:} \url{https://github.com/chrnx-dev/ailed-chess}

\textbf{Zenodo DOI:} \url{https://doi.org/10.5281/zenodo.18805494}

The codebase includes the complete signal chain implementation, psyche calculator, move selector, all experiment scripts, and reproduction instructions. The system is pure Python with no compiled dependencies, making it readily deployable and compatible with any chess engine that outputs move probabilities. Readers interested in integrating the signal chain into their own systems should refer to the \texttt{README.md} and \texttt{CITATION.cff} in the repository.

% ============================================================================
% REFERENCES
% ============================================================================

\end{document}